## REVIEW ARTICLE

# Evolutionary Reinforcement Learning: A Survey

Hui Bai[1], Ran Cheng[1]*, and Yaochu Jin[2,3]

[1]Department of Computer Science and Engineering, Southern University of Science and Technology, Shenzhen, China. [2]Faculty of Technology, Bielefeld University, 33615 Bielefeld, Germany. [3]Department of Computer Science, University of Surrey, Guildford, Surrey GU2 7XH, UK.

*Address correspondence to: ranchengcn@gmail.com

Reinforcement learning (RL) is a machine learning approach that trains agents to maximize cumulative rewards through interactions with environments. The integration of RL with deep learning has recently resulted in impressive achievements in a wide range of challenging tasks, including board games, arcade games, and robot control. Despite these successes, several critical challenges remain, such as brittle convergence properties caused by sensitive hyperparameters, difficulties in temporal credit assignment with long time horizons and sparse rewards, a lack of diverse exploration, particularly in continuous search space scenarios, challenges in credit assignment in multi-agent RL, and conflicting objectives for rewards. Evolutionary computation (EC), which maintains a population of learning agents, has demonstrated promising performance in addressing these limitations. This article presents a comprehensive survey of state-of-the-art methods for integrating EC into RL, referred to as evolutionary reinforcement learning (EvoRL). We categorize EvoRL methods according to key research areas in RL, including hyperparameter optimization, policy search, exploration, reward shaping, meta-RL, and multi-objective RL. We then discuss future research directions in terms of efficient methods, benchmarks, and scalable platforms. This survey serves as a resource for researchers and practitioners interested in the field of EvoRL, highlighting the important challenges and opportunities for future research. With the help of this survey, researchers and practitioners can develop more efficient methods and tailored benchmarks for EvoRL, further advancing this promising cross-disciplinary research field.

## Introduction

Reinforcement learning (RL) has achieved remarkable success in recent years, particularly with the integration of deep learning (DL) to solve complex sequential decision-making problems [1,2]. Despite these advancements, RL still faces several challenges, such as sensitivity to hyperparameters [3]; difficulties in credit assignment in tasks with long time horizons, sparse rewards, and multiple agents [4,5]; limited diverse exploration in tasks with deceptive rewards or continuous state and action spaces [6]; and conflicting objectives for rewards [7]. Moreover, numerous optimization problems in RL are typically complex black-box optimization problems with properties such as being gradient-free, non-convex, multi-modal, multi-objective, discrete, discontinuous, and dynamic [8,9]. However, traditional optimization methods, such as some signal processing optimization methods, struggle to solve these complex problems [10,11]. In contrast, evolutionary computation (EC) methods have successfully addressed these problems [12–15].

To address the aforementioned challenges and cater to the requirements of solving complex optimization problems in RL, the field of evolutionary reinforcement learning (EvoRL) has emerged by integrating RL with EC [16,17]. EvoRL involves maintaining a population of agents, which offers several benefits, such as providing redundant information for improved robustness [17], enabling diverse exploration [18], the ability to evaluate agents using an episodic fitness metric [17], and the ease of generating trade-off solutions through multi-objective EC algorithms [19].

EvoRL has a rich history, dating back to early work in neuroevolution, which used EC algorithms to generate the weights and/or topology of artificial neural networks (ANNs) for agent policies [20,21]. Since the proposal of OpenAI ES [16], EvoRL has gained increasing attention in both EC and RL communities. While there have been some surveys focusing on various aspects of EvoRL, such as neuroevolution [22], multi-objective RL [7,23], automated RL [24], and derivative-free RL [25], they either focus on a narrow research field within RL or lack a comprehensive overview of EC methods as applied to RL.

To bridge the gap between EC and RL communities, this article provides a comprehensive survey of EvoRL, elaborating on 6 key research fields of RL, as shown in Fig. 1. The EvoRL methods are introduced and discussed separately for each field, focusing on their advantages and limitations.

Finally, the article discusses potential improvement approaches for future research, including efficient methods in terms of EvoRL processes, tailored benchmarks, and scalable platforms.

## Background

### Reinforcement learning

RL is a powerful tool for decision-making in complex and stochastic environments. In RL, an agent interacts with its environment by taking a sequence of actions and receiving a sequence of rewards over time. The objective of the agent is to maximize the expected cumulative reward. This problem can be modeled as a Markov Decision Process (MDP), which is defined as $<S, A, T, R, \rho_0, \gamma>$, with a state space $S$, an action space $A$, a stochastic transition function $T : S \times A \rightarrow P(S)$ that represents the probability distribution over possible next states, a reward function $R : S \times A \rightarrow \mathbb{R}$, an initial state distribution $\rho_0 : S \rightarrow \mathbb{R}_{\in [0,1]}$, and a discount factor $\gamma \in [0, 1)$.

The agent's behavior is determined by its policy, which is denoted by $\pi_\theta : S \rightarrow P(A)$, with $P(A)$ being the set of probability measures on $A$ and $\theta \in \mathbb{R}^n$ being a vector of $n$ parameters. The

agent updates its policy over time to maximize the expected cumulative discounted reward, as given by

$$J(\pi) = \mathbb{E}_{\rho_0, \pi, T} \left[ \sum_{t=0}^{\infty} \gamma^t r_t \right] \tag{1}$$

where $s_0 \sim \rho_0(s_0)$, $a_t \sim \pi(s_t)$, $s_{t+1} \sim T(\cdot \mid s_t, a_t)$, and $r_t = R(s_t, a_t)$.

RL algorithms can be divided into 2 categories: model-based and model-free. While model-based algorithms establish a complete MDP by estimating the transition function and reward function, by contrast, model-free algorithms are data-driven and optimize the policy by using a large number of samples, without the need to know the transition function and reward function. Due to the difficulties of establishing a complete MDP and the success of neural networks (NNs) in representing policies, model-free RL has become the main focus of research in recent years [26]. In this survey, we focus on model-free RL methods.

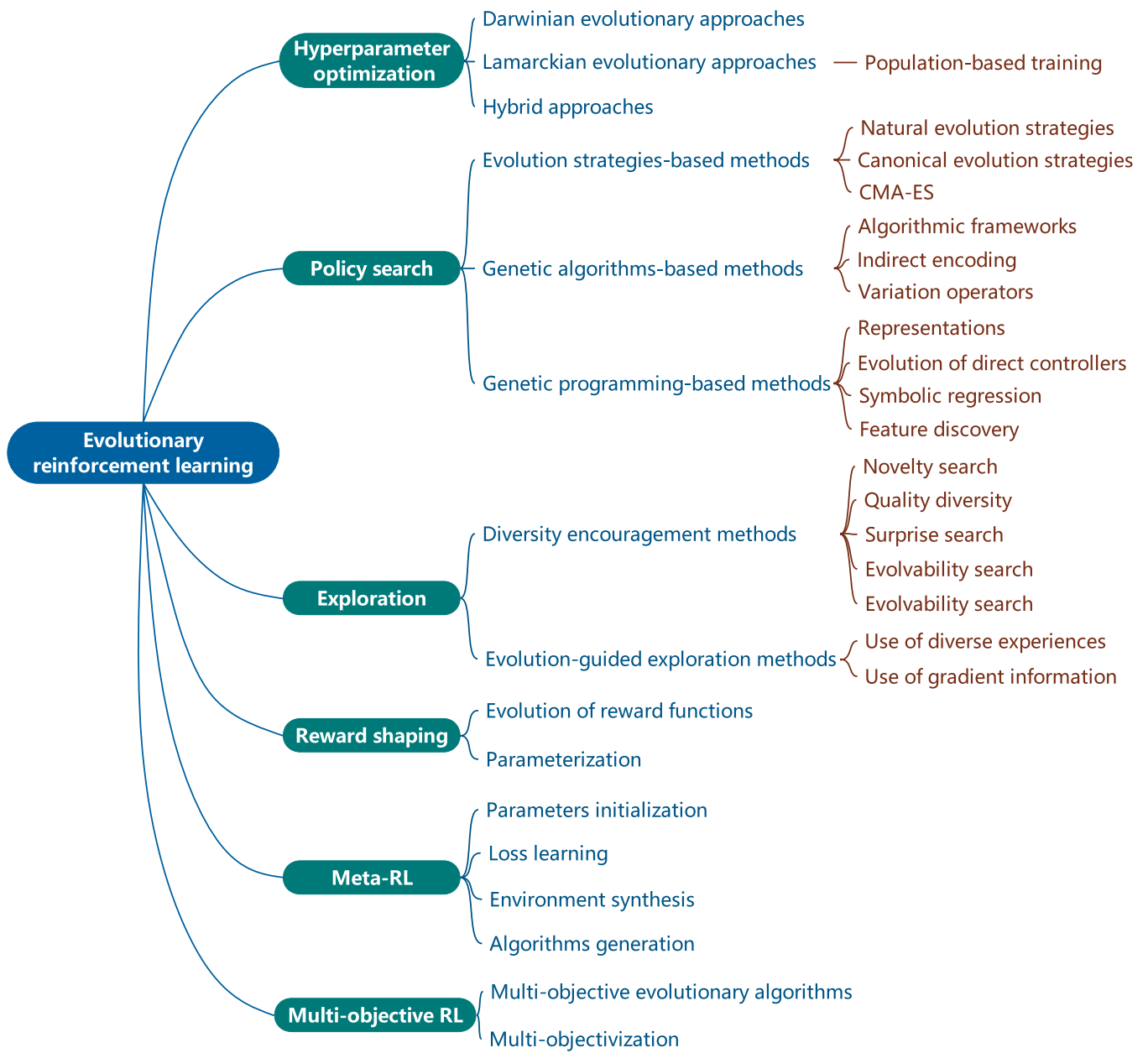

**Fig. 1.** Key research fields of evolutionary reinforcement learning. Hyperparameter optimization is a universal method for algorithms in the other 5 research fields to realize end-to-end learning and improve performance simultaneously. Policy search seeks to identify a policy that maximizes the cumulative reward for a given task. Exploration encourages agents to explore more states and actions and trains robust agents to better respond to dynamic changes in environments. Reward shaping is aimed at enhancing the original reward with additional shaping rewards for tasks with sparse rewards. Meta-RL seeks to develop a general-purpose learning algorithm that can adapt to different tasks. Multi-objective RL aims to obtain trade-off agents in tasks with a number of conflicting objectives.

More specifically, model-free RL methods can further be divided into 2 categories: policy-based and value-based methods. In policy-based methods, the parameters $\theta$ of the policy are adjusted in the direction of the performance gradient, according to the Policy Gradients Theorem [1]. Some of the state-of-the-art policy-based algorithms include Trust Region Policy Optimization (TRPO) [27], Proximal Policy Optimization (PPO) [28], Asynchronous Advantage Actor–Critic (A3C) [29], deep deterministic policy gradient (DDPG) [30], Twin Delayed DDPG (TD3) [31], and Soft Actor–Critic (SAC) [32]. In value-based methods, a parameterized $Q$-function is optimized to estimate the value of a state–action pair. One of the state-of-the-art value-based methods is Deep Q-Network (DQN) [33], which updates the parameters of the $Q$-function by minimizing the temporal difference (TD) loss using a batch of samples. Techniques such as experience replay [34] and double Q-network [35] have been proposed to improve the sample efficiency and exploration of DQN.

## Evolutionary computation

EC refers to a family of stochastic search algorithms that have been developed based on the principle of natural evolution. The primary objective of EC is to approximate global optima of optimization problems by iteratively performing a range of mechanisms, such as variation (i.e., crossover and mutation), evaluation, and selection. Among various EC paradigms, the evolution strategies (ESs) [36] are the mostly adopted in EvoRL, together with the classic genetic algorithm (GA) [37] and the genetic programming (GP) [38].

ESs primarily tackle continuous black-box optimization problems where the search space lies within the continuous domain. Therefore, it is predominantly applied to weight optimization of policy search in RL. ESs for RL are typically categorized into 3 main classes, which are canonical ES [39], covariance matrix adaptation ES (CMA-ES) [40], and natural ES (NES) [41]. Canonical ES is aimed at obtaining final solutions with high fitness values while using a small number of fitness evaluations through conducting an iterative process involving variation, evaluation, and selection. Here, we illustrate the canonical $(\mu, \lambda)$-ES algorithm. Following the initialization of policy parameters $x \in \mathbb{R}^n$ and a set of hyperparameters, the algorithm generates $\lambda$ offspring $x_1, \ldots, x_\lambda$ from a search distribution with mean $x$ and variance $\sigma^2 C$. Subsequently, all offspring are evaluated using a fitness evaluation function. Following this, a new population mean is generated by moving the old population mean towards the best $\mu$ offspring. Then, $\sigma$ is optionally updated, and an adaptive $\sigma$ can improve ESs performance. CMA-ES shares the same procedure with canonical ES but is more effective since its mutation step size $\sigma$ and covariance matrix $C$ are updated adaptively, enabling the capture of the anisotropy properties of general optimization problems. NES also shares the same iteration with canonical ES. However, it updates a search distribution iteratively by estimating a search gradient (i.e., the second-order gradient) on the distribution parameters towards higher expected fitness values. The distribution parameters are updated through estimating a natural gradient that can find a parameterization-independent ascent direction compared to a plain search gradient [42]. The natural gradient $\tilde{\nabla}_\theta J$ is formulated as $\mathbf{F}^{-1}\nabla_\theta J$, where $\mathbf{F}$ is the Fisher information matrix (FIM) of the parametric family of the search distribution, and $\nabla_\theta J$ is the estimated search gradient of expected fitness by Monte Carlo estimation. The FIM implies the degree of certainty of updating $\theta$, which has the effect of punishing the natural gradient with high variance and boosting the natural gradient with low variance.

GAs, as the most classic EC paradigm, are also commonly adopted in EvoRL. GAs follow the workflow where a population of candidate solutions is iteratively improved through selection, crossover, and mutation. The encoding or representation of the search space in GAs can be tailored to the specific problem at hand, allowing for binary encoding and discrete encoding for combinatorial optimization problems, and real encoding for numerical optimization problems. This versatility in encoding types makes GAs a widely applicable method for solving various problems in RL [43,44].

GP is a distinctive EC paradigm that is different from ESs or GAs, which mainly solve numerical optimization problems. In GP, the search space is composed of a set of programs that are represented using various encoding methods, such as abstract syntax tree, executable graph (e.g., Cartesian GP [CGP] and tangled program graphs [TPG]), finite-state machine, and context-free grammar, among others [36]. The fitness of a program is evaluated by executing it to observe its behavior, and the programs can be viewed as data when crossover and mutation operations are performed on them. In contrast, the data are interpreted as programs when they are executed. GP has several advantages over other EC paradigms, including the ability to handle complex problems that require a programmatic solution, such as symbolic regression and control problems, among others [45–47].

## Discussion

EC algorithms have been recognized as competitive tools to handle complex optimization problems that exhibit non-convex, non-differentiable, non-smooth, and multi-modal properties [48]. In RL, EC is particularly useful for complex problems with numerous local optima and is suitable for problems without gradient information. Furthermore, it is applicable for problems without explicit objective functions by novelty search (NS) [49]. The population-based search strategy of EC makes it robust to dynamic changes that are commonly found in real-world applications of RL, such as sim-to-real transfer in robot control [50]. A simple and general framework that combines EC and RL is shown in Fig. 2.

Specifically, EC has been introduced into RL in 6 major key research fields of RL, including hyperparameter optimization (HPO), policy search, exploration, reward shaping, meta-RL, and multi-objective RL, as presented in Fig. 1. Since the 3 EC methods (i.e., ESs, GAs, and GP) have differences in mechanisms, their target problems and focused applications to the 6 RL research fields are slightly different. The detailed comparisons are listed in Table. ESs employ real encoding and simulate gradient-based methods, and thus are especially applied to continuous optimization problems with a large number of decision variables (e.g., NN-based policy search). GAs adopt various kinds of encoding methods and involve diversity metrics, and therefore can be applied to various optimization problems in RL research fields. GP designs tree-based or graph-based encodings to generate interpretable and programmatic solutions and is mainly applied to regression problems or direct control problems. Next, these 6 key research fields of RL will be introduced in detail in the following sections, respectively.

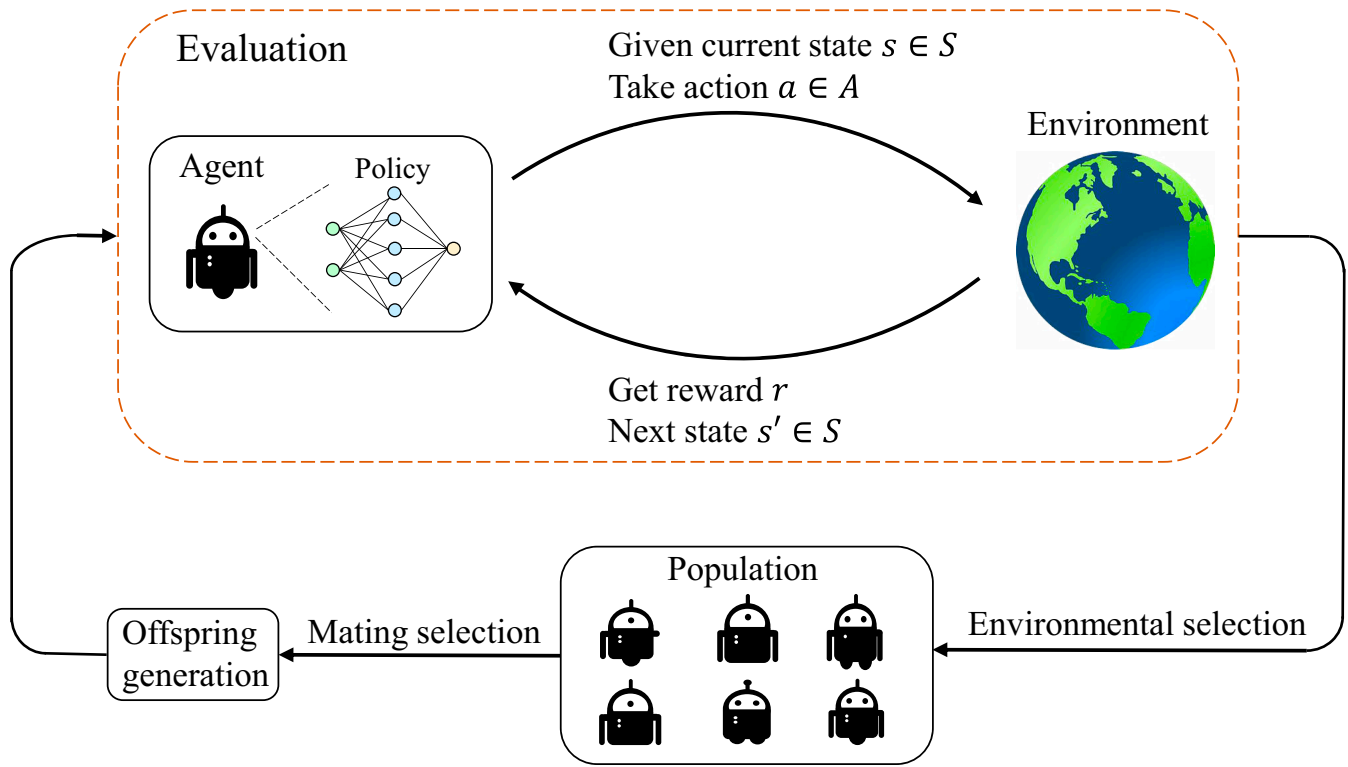

**Fig. 2.** A simple and general framework of EvoRL. The framework consists of 2 loops: the outer loop shows the evolution process of EC, while the inner loop illustrates the agent–environment interaction process in RL. Initially, a population of parent candidate solutions is randomly initialized, and then the offspring candidate solutions are generated from the parents via variation. Each offspring is evaluated using an RL task to obtain its fitness value, and a new population is selected for the next iteration by combining all parents and offspring.

**Table.** Comparisons of 3 types of EC paradigms (i.e., ESs, GAs, and GP) in terms of the differences in mechanisms, target problems in RL, and integration with RL research fields.

| EC methods | Differences in mechanisms | | | Target problems in RL | Integration with RL research fields |
|---|---|---|---|---|---|
| | Encodings | Variation | Selection criteria | | |
| ESs | Real encoding | Mutation | Performance metrics | Continuous optimization problems | Policy search, exploration, reward shaping, hyperparameter optimization, meta-RL, multi-objective RL |
| GAs | Integer encoding, real encoding, indirect encoding | Crossover, mutation | Performance and diversity metrics | General optimization problems | Policy search, exploration, reward shaping, hyperparameter optimization, meta-RL, multi-objective RL |
| GP | Syntax tree, executable graph | Crossover, mutation | Performance and interpretability metrics | Regression or control problems | Policy search, reward shaping, meta-RL |

## EC in HPO

Finding the optimal hyperparameter configurations for RL can be a challenging task due to the large number of hyperparameters involved, including those related to RL algorithms (e.g., the learning rate $\alpha$ and discount factor of future rewards $\gamma$) and the NN architectures of policies (e.g., the number and size of layers). To overcome this challenge, researchers have introduced HPO to automatically set configurations for optimal performance. HPO has been shown to improve the performance and robustness of RL algorithms [51,52].

However, HPO for RL faces several challenges. First, performance evaluation can be extremely expensive for complex tasks. Second, the search space of hyperparameters can be complex, involving mixed encoding, high dimensionality, and non-convexity. Third, there may be 2 or more objectives that need to be traded off. To address these challenges, there are several major classes of HPO methods, including random search [53], Bayesian optimization [54], gradient-based methods [55], and EC methods [51]. Among these, EC methods can simultaneously meet the challenges of HPO, owing to their high degree of parallelism, gradient-free properties, and ability to obtain a set of trade-off optimal solutions.

EC-based HPO methods can be mainly classified into 3 categories: Darwinian evolutionary methods, Lamarckian evolutionary methods, and hybrid methods. In Darwinian evolutionary methods, parameters are initialized while hyperparameters are evolved. In contrast, in Lamarckian evolutionary methods, parameters are inherited while hyperparameters are evolved. Additionally, hybrid methods further combine the former 2 methods and gradient-based methods.

### Darwinian evolutionary methods
In Darwinian evolutionary methods, the parameters are randomly initialized while hyperparameters are evolved using GAs [56]. For instance, Eriksson et al. applied GA to evolve 2 hyperparameters, the learning rate $\alpha$ and the temperature $\tau$, which control the trade-off between exploration and exploitation in softmax action selection, for Sarsa($\lambda$) in food capture tasks. Elfwing et al. [57] also applied GA to evolve hyperparameters and weights in potential-based reward shaping for Sarsa($\lambda$) in the same tasks. These methods integrate learning and evolution to effectively improve the performance of RL algorithms and obtain sim-to-real robust policies. Nonetheless, the Darwinian evolutionary methods are inefficient since parameters are reinitialized in each generation, causing a loss of knowledge already acquired during previous generations.

### Lamarckian evolutionary methods
In Lamarckian evolutionary methods, parameters are inherited while hyperparameters are evolved, meaning that hyperparameters are adapted to the current learning process to make agents learn more efficiently. A state-of-the-art asynchronous parallel evolutionary method called population-based training (PBT) has recently been proposed to improve the efficiency of HPO [51]. In the asynchronous PBT, only one ready individual is compared with a randomly selected individual from the remaining population in each generation, and then the worse individual copies the parameters and hyperparameters of the better individual and adds noise to its hyperparameters. PBT has successfully trained a series of RL agents in a number of complex tasks, such as the 3D multi-player first-person video game, DMLab, the MuJoCo multi-agent soccer game, and ELF OpenGo, achieving new state-of-the-art performance of RL algorithms [58–61]. The PBT-style evolution is quite similar to the steady-state EC (i.e., a new individual is inserted into the population in each generation) that is believed to be effective for non-stationary/dynamic environments [62]. However, PBT does not consider the diversity of the population. That is, PBT prefers the higher-performing configurations, but it may lose the individuals that are "late bloomers". Therefore, the faster improvement rate PBT (FIRE PBT) has been proposed based on the assumption that when 2 NNs have similar performance and hyperparameters, the NN with a faster rate of improvement will bring about a better final performance [63]. FIRE PBT derives a fitness metric based on the assumption and introduces subpopulations to increase diversity. Further, a sample-efficient automated RL framework (SEARL) has been proposed for off-policy RL algorithms, which follows the PBT style to evolve dynamic configurations of hyperparameters and shares experiences across the population with different configurations [52].

### Hybrid methods
Hybrid methods combine the above 2 evolutionary methods and other gradient-based methods to improve training efficiency. For example, in the work of Fernandez and Caarls [64], a hybrid method combines Darwinian and Lamarckian evolutionary methods, following the mainstream of Lamarckian evolution while conducting multiple random restarts of parameters in the evolutionary process to escape local optima. In addition, an evolutionary stochastic gradient descent (SGD) framework is proposed, aiming at combining the merits of SGD and EC [65]. In the framework, a set of NN weights with distinct hyperparameters are optimized independently by various SGD variants, and then their information is exchanged by EC. However, the initial hyperparameters are preset by humans, which still involves a certain amount of human knowledge. Moreover, Schneider et al. [66] have proposed a collection of benchmarks derived from HPO to verify the performance of quality diversity (QD) methods. In other words, HPO methods can be resolved by QD methods to introduce diversity by niches.

### Discussion
HPO is crucial for achieving state-of-the-art performance in RL, and EC methods have shown great potential in automating this process. However, the current literature on HPO still faces several challenges, such as the lack of comprehensive performance metrics and the need for efficient convergence speed. Additionally, selecting hyperparameters from a large number of options is a combinatorial optimization problem that brings new challenges to EC methods.

To address these challenges, future research should focus on developing comprehensive evaluation metrics that consider both effectiveness and efficiency. This can be achieved by benchmarking different HPO methods on a wide range of tasks and considering factors such as training time, convergence speed, and performance. Additionally, future research should explore new EC methods that can efficiently search high-dimensional and combinatorial search spaces for HPO. Furthermore, it is essential to investigate how to combine different HPO methods to improve the overall efficiency and effectiveness of the optimization process. By addressing these challenges, EC-based HPO can further accelerate the development and deployment of RL algorithms in real-world scenarios.

## EC in Policy Search

In the context of RL, policy search seeks to identify a policy that maximizes the cumulative reward for a given task. The incorporation of NNs as function approximators for policies has been facilitated by the surge of DL, despite the vast search space of states and actions. SGD methods are widely used for training NN weights in deep RL. Alternatively, neuroevolution has emerged as an alternative approach, leveraging gradient-free EC methods for policy search, which can optimize NN weights, architectures, hyperparameters, building blocks, and even learning rules [67].

Early work in neuroevolution focused primarily on the evolution of the weights of small and fixed-architecture NNs. Recent advancements, however, have demonstrated the promise of evolving the architecture together with the weights of NNs for complex RL tasks [21]. Moreover, a new perspective on policy search has been established by ignoring the weights and conducting only architecture search [68]. This section will review EC techniques such as ESs, GAs, and GP for policy search in RL.

### Evolutionary strategies-based methods

The subsection will review 3 popular ESs used in RL tasks, namely, canonical ES, NES, and CMA-ES.

#### Canonical ES-based methods

The high-parallel framework of OpenAI ES has led to the successful application of a simplified canonical ES to Atari games [69]. Prior to this, the canonical ES had not been applied much to RL tasks, since it performs poorly in high-dimensional tasks. Although the canonical ES can achieve similar performance to OpenAI ES on several Atari games with discrete state and action spaces, its performance on continuous state and action spaces (where EC is more preferable [70]) has not been investigated.

As ESs treat RL tasks as black-box optimization problems directly instead of taking advantage of their intrinsic MDP structures, it may have a large variance in multiple runs. To address this issue, several variance reduction techniques have been introduced. On the one hand, various ESs gradient estimators using Monte Carlo techniques have been proposed, such as the antithetic ES gradient estimator in OpenAI ES [16] and the forward finite-difference ES gradient estimator in the structured ES [71]. Specifically, the structured ES performed well on most of the MuJoCo tasks using less than 300 policy parameters. Furthermore, the adaptive ES-active subspace method further combines structured ES with the techniques from active subspaces to learn the changing dimensionality of the gradient space, which achieved competitive performance compared to PPO, TRPO, and several ES variants on a subset of MuJoCo tasks [72]. On the other hand, structured methods leveraging the underlying MDP structures have been developed. Specifically, the control variate, also known as the advantages function in RL [29], has been introduced into ESs to reduce the variance of Monte Carlo gradient estimation [73].

Since applying ESs to large-scale RL tasks is low-efficient, a number of works have been proposed to improve sample efficiency in 2 ways: sampling from diverse search directions and making full use of previous samples. In the first way, the Gaussian orthogonal exploration searches a number of diverse directions [71]. Based on this method, the guided ES further combines ES with surrogate gradients (correlated with the true gradients) [74]. Then, the self-guided ES trades off exploitation in the gradient subspace and exploration in its orthogonal complement subspace, which has obtained higher returns and faster convergence speed than PPO, TRPO, and guided ES on several MuJoCo tasks [75]. In the second way, the trust region ES (TRES) approximately optimizes a surrogate objective by reusing the data sampled from the old policy parameters instead of sampling from new parameters, which has achieved faster convergence speed than PPO and TRPO on several MuJoCo tasks [76].

#### NES-based methods

The original NES is known to be limited in scalability for high-dimensional problems due to the time-consuming calculation of FIM [41]. To improve efficiency and robustness, the exact NES computes the inverse of the exact FIM instead of the empirical FIM and has shown competitive performance on the double pole balancing task using an NN with 21 weights [77]. However, NES was not successful for large-scale tasks until the development of OpenAI ES [16], which is a simplified variant of NES with an isotropic multivariate Gaussian and fixed variances $\Sigma$. OpenAI ES is closely related to the policy-based RL algorithm PEPG in the theoretical relation [78]. To reduce variance and realize high-parallel ability, OpenAI ES has introduced several techniques such as mirrored sampling of the perturbation, rank-based fitness shaping, virtual batch normalization, and the random seeds sharing strategy. As a result, OpenAI ES has achieved competitive performance on MuJoCo and Atari games with over a million policy parameters by using thousands of central processing unit (CPU) workers, demonstrating the advantages of ESs as a black-box optimization algorithm for complex RL tasks. Moreover, OpenAI ES has been shown to be close to SGD with a large number of offspring [79] and resembles traditional finite-difference approximators [80].

Several diversity encouragement methods from EC such as NS and QD have been introduced to enhance the exploration of OpenAI ES, encouraging agents to exhibit diverse behaviors [6]. Hybrid algorithms NS-ES and NSR-ES can solve tasks with noisy and deceptive rewards. Furthermore, progressive episode lengths (PELs) have been proposed to improve the learning efficiency in evaluating fitness values of samples [81]. PEL enables agents to learn from simple tasks to complex tasks by dividing the time budget and episode lengths into increasing numbers of fragments concurrently.

#### CMA-ES-based methods

The application of the CMA-ES to RL was first proposed by Igel in 2003, who demonstrated that CMA-ES can achieve faster convergence than several state-of-the-art GA-based neuroevolution algorithms on double pole balancing tasks using a single hidden layer policy [82]. CMA-ES typically uses rank-based fitness shaping instead of absolute fitness values to reduce its susceptibility to noise. However, the accuracy of rank still plays a critical role in performance. To address this, Heidrich-Meisner augmented CMA-ES with Hoeffding- and Bernstein-based racing algorithms to obtain a reliable rank, which led to faster convergence and more robust hyperparameter selection on single and double pole balancing tasks than several GA-based neuroevolutionary algorithms [83,84].

Recently, Chen proposed a restart-based rank-1 ES (R-R1-ES), a simplified CMA-ES, to play Atari games using a 2-hidden-layer

NN, which is a groundbreaking work applying efficient CMA-ES to complex RL tasks [85]. R-R1-ES integrates a Gaussian-distributed model with 2 mechanisms, including the adaptation of the number of parents and a restart procedure, and has achieved higher scores than OpenAI ES, canonical ES, NS-ES, and NSR-ES [88] on a subset of Atari games. Despite the development of several high-efficient CMA-ES variants, such as R1/Rm-ES [86], LM-MA-ES [87], and fast CMA-ES [88] to deal with the time-consuming adaptation of the full covariance matrix for large-scale optimization problems with up to 10,000 dimensions, OpenAI ES has been verified on millions of policy parameters. Therefore, their potential for complex RL tasks is yet to be studied.

## GA-based methods

GAs have been widely adopted to optimize the weights and architectures for policy search in RL, owing to their diverse encoding types. The GA-based methods have primarily focused on 3 research topics: algorithmic frameworks, indirect encoding, and variation operators.

### Algorithmic frameworks

**Pure GA-based frameworks.** In the 1990s, several studies utilized GAs to optimize policy weights for pole balancing problems [20,89]. Among these works, GENetic ImplemenTOR (GENITOR), which represented weight with real values instead of binary strings, improved the precision and efficiency of the search space. Subsequently, a range of works has been developed, with the most popular one being NeuroEvolution of Augmenting Topology (NEAT), which can obtain a minimal NN by adding nodes and connections from the smallest NN without hidden nodes [21]. NEAT has several highlights, including genetic encoding that aligns corresponding genes easily when mating 2 genomes, historical markings that enable tracking and matching of genes during crossover, and speciation within smaller niches to protect topological innovations. NEAT has since been improved and tailored to various tasks, such as evolving dynamic policies to adapt to environmental changes in a dangerous foraging task [90], evolving complex policy architectures in robot competition and coevolution tasks [91], video games [92], and strategic decision-making problems [93].

Despite the effectiveness of NEAT, it does not fully optimize the weights under a potential architecture. Hence, ENAT was developed based on NEAT, which adopts the idea of incremental growth from a minimal structure [94]. However, ENAT applies CMA-ES to fully optimize the weights and introduces a compact genetic encoding to encode a tree-based program in a linear genome. As a result, ENAT can find better weights than NEAT with the same network size on a robot arm control task. To reduce the side effect of topology change, CMA-TWEANN replaces the mutation by random weights in NEAT with a seamless topology mutation by zero weights and applies CMA-ES to optimize weights [95]. However, whether weight optimization is more important than topology optimization is still debatable. The weight-agnostic search method answers this problem by only searching the topology by NEAT without training of weights. This method has found policies with minimal architectures in continuous control tasks [68].

In the highly parallel framework of OpenAI ES, a simple GA has been used to optimize large-scale policies with millions of weights in Atari games and MuJoCo [96]. In a similar vein, a massively parallel method has been applied to search recurrent neural network (RNN) architectures using only mutation. This method has achieved high performance with orders of magnitude fewer parameters than several state-of-the-art RL methods in MuJoCo tasks [97]. While these methods require huge computational resources, the training of large-scale policies using EC methods is still very low-efficient. To improve the efficiency, a hybrid agent model consisting of a large world model and a small controller model has been proposed to tackle complex RL tasks [98]. The world model extracts low-dimensional features from real-world observations and predicts future states based on historical information. The controller model, such as a single-layer linear NN, determines the actions to take by receiving current and predicted features to maximize the expected cumulative reward. The controller model has been evolved by EC methods in vision-based game tasks [99,100]. Moreover, an end-to-end training of the whole agent model using GAs has shown comparable performance in car racing tasks [101].

**Frameworks hybridizing GAs and RL.** A combination of GA-based methods (i.e., NEAT) and TD methods (i.e., Q-learning) has led to the development of 2 methods, namely, Lamarckian NEAT+Q and Darwinian NEAT+Q. In these methods, NEAT is used to optimize the architectures and initial weights of Q networks, while the policy weights are updated using backpropagation [102]. To balance the exploration and exploitation of the EC method, $\varepsilon$-greedy selection and softmax selection in RL have been incorporated into NEAT. However, the high sample complexity of NEAT+Q, resulting from the fully training of each candidate policy in highly stochastic domains, has prompted the proposal of an efficient NEAT+Q method that reuses previous samples to train a population of candidate policies [103]. A comparison study has suggested that EC methods are more effective when fitness can be rapidly evaluated in deterministic domains, whereas TD methods exhibit an advantage in fully observable but non-deterministic domains [104].

**Cooperative coevolution.** In the field of NN optimization, the search space can become excessively large when the numbers of the NN input, output, architecture, and weight are large. To overcome this challenge, cooperative coevolution (CC)-based GAs can be used, which decompose the problem into smaller components to reduce complexity and enable more efficient resolution through CC methods [105]. In a CC algorithm, each individual represents a partial solution, or a component of a complete solution, which is resolved by a species, or a set of individuals, independently. The individuals are evaluated based on their contributions to the complete solution. This approach introduces diversity and robustness in the maintenance of various components and enables parallel search to improve training efficiency. The search granularity in the decomposition methods of NNs indirectly influences search performance and efficiency. In neuron-level CC methods, weights connected to a neuron are grouped into a component. Symbiotic adaptive neuro-evolution (SANE) is an earlier CC algorithm that evolves a population of hidden neurons for an NN with a fixed architecture, and has shown better performance than Q-learning and GENITOR on pole balancing tasks [106]. Enforced Sub-populations (ESP) improves the efficiency of SANE and supports the evolution of RNNs by allocating a species for a hidden neuron and conducting variation within species [107]. NSP further groups the weights connected to a hidden neuron into finer granularity [108]. In synapse-level CC methods, each weight is considered a component. Based on ESP, CoSyNE groups each weight into a component and has shown better efficiency than SANE, NEAT, and ESP on pole balancing tasks [109]. However, CoSyNE may not be applicable for large-scale NNs since it cannot fully exploit the weights to avoid inaccurate evaluation, which is a problem in neuron-level methods. Other state-of-the-art decomposition methods include COVNET [110], Modular NEAT [111], and CCNCS [112].

### Indirect encoding
Research on indirect encoding has been promoted for 2 reasons. Firstly, direct encoding has limitations in scaling up to large-scale NN scenarios. Secondly, in biological genetic encoding, phenotypes typically contain more genetic components than genotypes, and the mapping of phenotypes to genotypes is indirect.

Artificial embryogeny, which includes cellular encoding [113] and generative encoding [114], utilizes a developmental phase by reusing genes or rules to evolve artificial systems from a small starting point [115]. In addition to repetition by reuse, the properties of physical space, such as symmetry and locality, have motivated the design of encoding that seeks to discover regularity by local connectivity. Compositional pattern-producing networks (CPPNs) capture structural relationships and are encoded by a composition of functions organized in the form of NNs [116]. CPPNs can be trained in the same way as NNs, and HyperNEAT is tailored to training CPPNs, which can evolve increasingly complex expression patterns to capture the complete regularities of problem structures [117,118]. The ability to learn from geometric regularity has enabled HyperNEAT to be successfully applied to complex tasks such as checkers [119], Go [43], and Atari games [120]. Additionally, several HyperNEAT variants have been proposed, such as adaptive HyperNEAT [121] and ES-HyperNEAT [122].

By modeling modularity as an optimization objective, NSGA-II (a multi-objective evolutionary algorithm) [123] has been applied to evolve CPPNs. This method performs better than HyperNEAT by generating lower modularity of genotypes and phenotypes in a robotics task [124]. However, CPPNs may lose continuity when mapping genotypes to phenotypes (i.e., a small change in the genotype may lead to a large change in the phenotype). Therefore, compressed encoding uses discrete cosine transform (DCT) to reduce the dimensionality of the search space by exploiting the spatial regularities of the weight matrix and obtains large-scale NNs on a vision-input car driving task [125].

Furthermore, several works have combined indirect encoding and direct encoding to discover regularities by indirect encoding and compensate for irregularities by direct encoding [126–128].

### Variation operators
Variation operators aim to preserve the characteristics of parents while introducing diversity, and they typically include crossover and mutation. Crossover combines the properties of more than one parent, while mutation inherits the properties from one parent to a large extent. In earlier binary encoding, the single-point crossover and flip mutation are widely used operators [36]. Since the length of the binary string determines the representation precision of a real number and influences the variation granularity, real encoding has been proposed to represent a weight as a real number directly. Accordingly, the simulated binary crossover and the polynomial mutation have been proposed for continuous search space [129]. Gaussian mutation is also widely used by adding a random value from a Gaussian distribution to a real-encoded weight [96].

However, since NNs are sensitive to small modifications of weights, the above variation operators typically cause catastrophic forgetting of the characteristics of parents. Hence, imitation learning or network distillation has been applied to variation operators for NNs, such as the state-space crossover [130], the Q-filtered distillation crossover [131], as well as the distilled topology mutation [132]. Both crossover methods apply imitation learning to distill better behaviors of parents into offspring. The mutation method first generates an offspring by augmenting topology and then pretrains the offspring by distilling the behavior of its parent as a necessary initialization.

Furthermore, Lehman et al. [133] propose a family of safe mutation to deal with the catastrophic forgetting problem, where the mutation degree of each weight is scaled by the sensitivity of the weight to the NN outputs, and thus an offspring will not diverge too much from its parent. In contrast, Marchesini et al. [134] propose a different concept of safe mutation for safe exploration, which uses visited unsafe states to explore safer actions.

Apart from the catastrophic forgetting problem, the permutation problem (i.e., the same solution can be represented by different NNs) can lead to low-efficient search when encoding schemes or variation operators are not designed sophisticated [135]. Hence, 2 types of approaches are developed. One approach tailors encoding methods for evolving NN architectures. For example, NEAT designs a genetic encoding to track parent solutions by using an innovation number [21]. Another approach aligns neurons from 2 NNs by analyzing their functional correlations before crossover and provides a general method for different encoding [135].

## GP-based methods
GP is a popular method for solving complex RL tasks since a program can emulate any model of computation given sufficient time and search space [136]. Three commonly used representations in GP are the abstract syntax tree, CGP, and TPG. These representations are applied to various tasks such as evolving direct controllers, symbolic regression, and feature discovery for RL tasks.

### Representations
The syntax tree is a commonly used representation in GP, where programs are directly encoded as genomes. As shown in Fig. 3A, nodes can represent functions, operations, variables, and constants, and the tree outputs a unique program through tree traversal algorithms [36]. Typically, the tree grows randomly by crossover and mutation from a null node, but it can suffer from the bloat issue, where programs grow in size without showing obvious fitness improvement over time if depth limiting is not enforced.

To address the bloat issue, CGP uses an integer-array genome with a fixed length to encode an executable tree graph [137], as depicted in Fig. 3B. All nodes are placed in a grid manner and are indexed sequentially. The nodes in the first column are input nodes. The integer-array genome is divided into blocks, with each block consisting of 3 integer genes representing a single non-input node, where the third integer specifies a function,

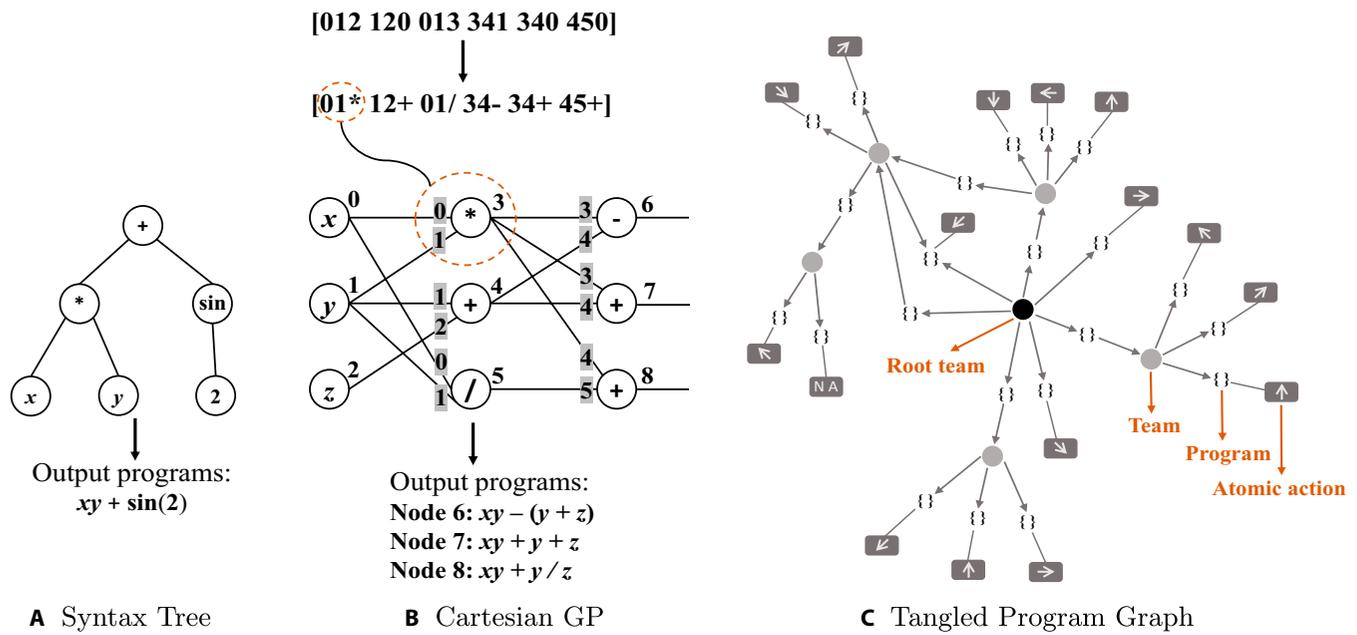

**A** Syntax Tree     **B** Cartesian GP     **C** Tangled Program Graph

**Fig. 3.** Three illustrative examples for the representations of GP in RL: (A) syntax tree, (B) Cartesian GP (CGP), and (C) tangled program graphs (TGP), respectively.

and the former 2 integers specify the indices of its input nodes. The data flow is from left to right in the graph, and the graph can have multiple output programs. Finally, users can specify only one output among the multiple outputs.

TPG is a framework for organizing multiple programs into a structure with high modularity [138]. TPG evolves 2 populations: a Node (i.e., Team) population and a Program population, where the Node population constructs a good organization of multiple teams of programs from the Program population, and the Program population discovers programs that output useful Atomic Actions. As shown in Fig. 3C, the black point represents the root node that receives state inputs. Evaluation of an agent starts at the root node and reaches an Atomic Action through a path in the direction of the arrow. This process only executes a fraction of the programs for a task, making TPG more efficiently than other neuroevolutionary algorithms that require covering all topology for a decision.

### Evolution of direct controllers

The syntax tree directly corresponds to the parse tree created by compilers, and has been applied to evolve controllers for robot control tasks [139], bipedal locomotion tasks [140], acrobat tasks, and helicopter hovering tasks [141] through syntax tree-based GP. Integration of the tree-based GP and RL has been tailored for real robots such that a precise simulator is not required in complex robot control tasks [142]. This method executes GP in a simplified simulator to generate simple controllers, which are then adapted to a particular real robot by RL. It has outperformed Q-learning on 2 complex robot tasks. In addition, cellular encoding and the syntax tree have been integrated to evolve NNs with particular structures [143]. Furthermore, a hybrid of NEAT and GP, known as HyperGP, has been applied to evolve weights of CPPN, and showed similar performance to HyperNEAT with significantly fewer evaluations [45].

Both CGP and NNs can be viewed as executable graphs, which has motivated researchers to extend the flexible CGP representation to the evolution of NNs. The CGP-based NN has been proposed to evolve both topology and weights for NNs with feed-forward or recurrent architectures, outperforming NEAT, ESP, and CoSyNE in pole balancing tasks [144]. Additionally, CGP has been applied to evolving the parameters of transfer functions such as the Gaussian function and the logistic sigmoid function, which improved the performance of NNs on a ball throwing task [145]. Furthermore, CGP has been applied to evolving game-playing agents directly from high-dimensional pixel inputs of Atari games [146], and the evolved programs are easier to understand than NNs.

TPG is tailored for visual RL tasks with high-dimensional pixel state inputs from environments, and has been applied to 20 challenging Atari games, where TPG exceeds DQN in 15 of the 20 games, and further exceeds the human level in 7 of the 15 games [147]. Surprisingly, TPG requires significantly lower computational resources and is without specialized hardware such as graphics processing units (GPUs). Besides, by interacting with environments, TPG introduces emergent modularity and thereby leads to task decomposition. Due to these advantages, TPG is capable of generating multi-task policies in Atari games [148,149], and has been successfully applied to the partially observable ViZDoom [150] and Dota2 [151].

### Symbolic regression

Interpretable RL is of high interest to academics and industrial areas, and interpretable controllers are more likely to be employed especially in industrial systems. The symbolic regression of GP is an effective approach to interpretable RL by fitting the policy or value functions in a human-understandable way. The syntax tree is naturally the appropriate representation due to its high interpretability and understandability for humans. For example, the value function discovery method has been proposed to discover algebraic expressions of an obtained

V-function by minimizing a simulation error between the expressions and sampled V-function data [152]; the GP for RL method has been proposed to generate smooth algebraic policies by the data sampled from the world models [153].

In addition to interpretability, symbolic regression can generate smoother and more adaptive symbolic approximators than numerical approximators, such as NNs. For example, a variant of the single-node GP has been applied to evolve a smooth proxy of the V-function by maximizing the number of correct choices of actions for sampled training states [154]. In addition, the single-node GP has constructed a symbolic process model for model-based RL methods to reduce the number of training data and adapt to the dynamic system in real-time robot control tasks. This method has advantages over NNs in that it requires no hyperparameter tuning and can generate smooth V-functions [155].

### Feature discovery

Feature discovery is the process of transforming input data into a form that can be more easily processed by RL agents. Unlike most feature extraction methods in machine learning, where useful features are extracted from inputs, feature discovery in RL may add features to aid in better learning by agents. For example, in pole balancing tasks, the optimal policy can be learned faster by adding 2 angle features of the pole.

In the works of Girgin and Preux [156] and Krawiec [157], each individual encodes programs with multiple S-expressions (feature functions), where each S-expression corresponds to a unique feature. In Ref. [157], the number of useful features is fixed, while in Ref. [156], the number of useful features can vary within a range since the prior is unknown in advance. The obtained features are human-readable, allowing for fine-tuning and knowledge transfer during the process of feature discovery.

### Discussion

In general, EC-based methods have shown great potential in solving complex RL tasks. Each method has its unique advantages and disadvantages, making them suitable for different situations. ESs are simple and efficient but can be limited by the high-dimensional search space and fitness noise. NES improves upon ES by using the reward gradient of all offspring, but there is still room for better exploitation of low-fitness offspring. GA-based methods can employ safe mutation methods to broaden their applicability. GP-based methods, such as CGP, can evolve policies with better interpretability, and TPG is a unique method that is specifically tailored to visual RL tasks and can solve challenging games with high-dimensional pixel inputs while using fewer computational resources than other EC based methods.

It is also worth noting that there are still areas for improvement in EC-based RL methods. The highly parallel framework of OpenAI ES requires a large number of CPU resources, which is low-efficient for large-scale image inputs, and allocating more resources to promising samples may enable better performance within limited computational resources. Furthermore, ENAS has shown great potential in automatically designing the architecture of deep NNs for image classification, but more research is needed to explore its applicability to policy search of complex RL tasks. Overall, as EC-based RL methods continue to evolve and improve, their potential in solving complex RL tasks makes them an exciting area of research.

## EC in Exploration

In RL, agents must interact with environments by taking actions and observing environmental states to collect trajectories to improve their behaviors. The learning efficiency of agents relies on the data that they gather. However, if an agent only visits a small portion of its environment, its knowledge will be limited, leading to suboptimal decision-making. Therefore, diverse exploration of the environment is desired. Agents typically explore environments by adding noise to the action space or to the parameter space of their policies. In state-of-the-art methods, the $\epsilon$-greedy exploration method encourages agents to take different feasible actions instead of the current optimal action for a state with a certain probability [1], while the parameter space noise method adds Gaussian noise to the policy weights to change the original output actions [158].

To achieve efficient exploration, 4 key challenges must be addressed [159]. First, the state–action space is often large, making it challenging for the agent to access the effective space. Second, the environment returns sparse and delayed rewards, meaning that agents cannot receive timely and informative feedback on their behaviors. Third, real-world environments often contain highly random and unpredictable things (i.e., whitenoise problems), making it challenging for the agent to distinguish important information from unimportant information, leading to unstable and inefficient exploration. Fourth, exploration in multi-agent RL is more challenging since the state–action space increases exponentially, and agents must explore coordinately to achieve local and global exploration trade-offs.

To deal with these challenges, EC methods for RL enable extreme exploration, competition, and cooperation, and massive parallelization by maintaining a set of diverse agents during search. In the implementation, EC methods introduce exploration for RL from 2 perspectives: diversity encouragement methods, especially for neuroevolution, and evolution-guided exploration methods for traditional RL algorithms.

### Diversity encouragement methods

In diversity encouragement methods for neuroevolution, EC as a policy search method directly evolves a set of diverse agents by encouraging diversity in the parameter space or in the behavior space. Most initial work focuses on the parameter space with the purpose of avoiding local optima. The widely applied diversity maintenance techniques have speciation (i.e., niching) and fitness sharing [21]. Speciation divides a population into a number of species according to their genetic similarity, and fitness sharing enables individuals with similar genomes to share their fitness so that innovation can be protected in their own species. However, diversity in the parameter space cannot ensure diversity in the behavior space, since there are infinite ways for NN weight settings to produce the same outputs of behaviors. Therefore, a number of recent approaches from or inspired by the diversity maintenance techniques of EC have been introduced into RL to directly reward diverse behaviors or novel states, such as NS [49], QD [160], surprise search [161], evolvability search [162], and curiosity search [163].

### Novelty search

NS abandons the fitness objective while rewarding novel behaviors that are different from previous behaviors. Behavioral characterizations (BCs) should be first designed to map the high-dimensional search space into a lower-dimensional behavior space. Then, to measure the novelty of a newly generated individual, a novel metric is defined as the task-specific distance between behaviors. After that, NS can be easily integrated into EC algorithms with little change of replacing the fitness objective with the novel metric. NS has been applied to NEAT and outperformed fitness-based methods on the deceptive T-Maze and biped walking tasks [49,164]. Empirical studies demonstrated that NS can bring unique advantages over fitness-based EC methods in overriding the deceptiveness of most fitness functions and making the evolutionary process more open-ended.

However, when faced with a task with a large state–action space, pursuing novelty alone does not perform better than fitness-based methods [165]. Thus, in high-dimensional evolutionary robotics tasks, NS is used to augment fitness-based EC methods as a diversity maintenance technique or served as the second objective to be optimized simultaneously with the fitness objective [166]. NS is augmented with local competition (NSLC) to create a set of diverse locomotion evolutionary robotics [167]. Additionally, NS is combined with OpenAI ES in 3 ways. The first version, NS-ES, replaces the gradients of expected rewards with the gradients of expected novelty, and the other 2 versions, NSR-ES and NSRA-ES, trade off the gradients of expected rewards and novelty [6]. Empirical studies have shown that the 3 versions performed better than OpenAI ES on Humanoid Locomotion and Atari tasks with deceptive traps. NS is also combined with sub-population to promote directed exploration in the population-guided NS method [168]. In general, all the above empirical studies have demonstrated that increasing behavioral diversity makes problems more easily resolved.

### Quality diversity

The pursuit of both fitness and novelty has led to the development of QD algorithms, which aim to find a large set of both diverse and high-performing solutions in a single run. The set of solutions aims to cover as many solution types, or BCs, as possible and find the best solution for each type. Two main state-of-the-art QD algorithms are NS with local competition (NSLC) and the multi-dimensional archive of phenotypic elites (MAP-Elites) [169]. A comparative study of the 2 algorithms in a set of maze tasks has revealed that the selection of BCs is a crucial and challenging issue, since it is task-dependent and should align with quality; otherwise, it can change the difficulty of finding a good solution [160]. Therefore, a number of automated BCs methods have been proposed to improve exploration efficiency, such as using dimensionality reduction methods to autonomously learn BCs [170,171], or mapping the high-dimensional parameter space into a low-dimensional manifold in which a high-density of good policies is located [172].

Moreover, exploration efficiency can be improved from other aspects. Several efficient behavioral diversity measurement methods have been proposed to measure the diversity of the entire population by determinants of behavioral embedding of policies [173] or use a string edit metric to measure behavioral distance [174]. To improve evaluation efficiency, the quality and novelty of new candidate solutions are predicted by a NN in open-ended robot object manipulation tasks [175]. To improve sample efficiency, a few-shot quality-diversity optimization method learns a population of prior policies for the initialization of QD [176]. To improve selection efficiency, an evolutionary diversity optimization algorithm with clustering-based selection selects a high-quality policy in each cluster for reproduction [177].

In addition, QD offers the potential for open-ended innovation that RL agents can generate and learn their own never-ending curriculum without human intervention. The paired open-ended trailblazer (POET) algorithm has been proposed to generate increasingly complex environments and optimize their solutions concurrently by combining the methods of NS, MAP-Elites, and minimal criterion coevolution [178,179]. The empirical studies on 2-D bipedal-walking obstacle-course tasks have demonstrated that solutions found by POET for challenging environments cannot be found by directly learning from scratch for the same environments. Further, a sample-efficient QD environment generation algorithm is proposed to apply a deep surrogate model to predict behaviors of agents in new environments [180]. Surprisingly, the open-ended coevolution of environments and solutions has provided novel ideas for addressing complex tasks.

### MAP-Elites

The focus of QD algorithms is mainly on MAP-Elites, which divides the behavior space of BCs into discrete bins according to the number of discretizations required for each dimension. Each bin records the best-found solution, and only one solution replaces a previous one if it outperforms the previous one in terms of both quality and diversity. MAP-Elites has been applied to generate elites of diverse behaviors (e.g., walking strategies) to help a robot adapt quickly to various kinds of damages [18], and has achieved better performance and robustness than PPO for simulated hexapod robot tasks [181]. However, MAP-Elites suffers from a scaling-up limitation in that the dimensionality of BCs must be low since the number of discrete bins increases exponentially with the dimensionality of BCs. To address this limitation, CVT-MAP-Elites uses centroidal Voronoi tessellation instead of grid-shaped bins to divide the behavior space into a desired number of regions [182]. In addition, several improvement methods have been proposed to scale up MAP-Elites to high-dimensional tasks. These include biased cell sampling [183] and gradient-based mutation operators [184] for efficient reproduction, approximated gradient [185], policy gradient-assisted MAP-Elites [186], and deep surrogate-assisted MAP-Elites [187] for the acceleration of optimization.

When facing hard-exploration tasks with sparse and deceptive rewards, RL algorithms, even with intrinsic motivation, perform poorly due to 2 challenges: detachment and derailment. Contemporary RL algorithms do not remember well-explored states (detachment), and random exploration may not lead back to well-explored states (derailment). Hence, Ecoffet et al. [4,188] proposed Go-Explore, a family of QD algorithms based on MAP-Elites. Go-Explore follows the key ideas of remembering states, returning to them (GO), and exploring from them (Explore). Go-Explore has greatly surpassed state-of-the-art RL algorithms on 2 challenging games such as Montezuma's Revenge and Pitfall.

### Surprise search

Surprise search is a new method of evolutionary divergent search that rewards deviation from the expected solution [161,189]. In contrast, NS rewards deviation from the prior solutions. Surprise search models the prediction of expected behavior and derivation. The expectation is based on the reasoning about past information, and thus surprise search can be viewed as a temporal novelty process. Both surprise search and NS are divergent novelty variants of QD, and their combination along with local competition has led to comparable fitness, higher efficiency, and better robustness (i.e., exploration and behavioral diversity) than NS on 60 highly deceptive maze navigation tasks [189].

### Evolvability search

Evolvability search is a new class of EC algorithms where the fitness function is a direct measure of the evolvability of an individual [162]. Evolvability is the potential for the future evolution of an individual. Evolvability search calculates the behavioral diversity of immediate offspring of an individual to estimate its future potential for diversity and then directly selects individuals with better potential diversity to enter into the next environmental selection. Encouraging behavioral diversity increases the adaptive ability of a lineage. Though resembling diversity-seeking methods such as NS, evolvability search outperforms NS on maze navigation and biped locomotion tasks [162]. However, evolvability search is computationally expensive due to its fitness evaluation process.

### Curiosity search

Curiosity search is a class of intrinsic motivated methods that learn intrinsic reward signals to enable agents to explore their environments. Exploration by intrinsic curiosity is a widely used method in RL algorithms, where intrinsic curiosity is used to complement the extrinsic rewards [190], predict the sequences of future actions or states [163], and achieve self-generated goals [191,192]. Goal exploration processes (GEPs) from EC methods explore robustly, designing a set of behavioral features (i.e., goals) based on the outcome trajectories of policies, and then exploit around these generated goals by directed behavioral diversity without being aware of external rewards [194]. GEP has been integrated with off-policy RL methods to exploit policy parameters, and the diverse samples generated by GEP can be inserted into the replay buffer of DDPG for training [192]. The intrinsically motivated GEP method integrates curiosity search and GEP to discover and acquire skills by self-generation, self-selection, self-ordering, and self-experimentation of learning goals [191]. Additionally, Stanton and Clune [193] reward intra-life novelty to encourage agents to explore new states within their lifetime. This method discretizes the pixel space into curiosity grids and rewards agents for visiting new locations on the grids. In contrast to the across-training novelty of NS, curiosity search can revisit previously visited potential states.

### Evolution-guided exploration methods

In evolution-guided exploration methods, the EvoRL method was the pioneer [17]. Since then, a number of works sharing the framework of ERL have been proposed. In these methods, EC introduces exploration in mainly 2 ways: EC agents generate diverse experiences stored in the replay buffer for the training of off-policy RL methods (e.g., DDPG, TD3, and SAC), or RL agents are directly updated using the gradient information of EC agents.

### Use of diverse experiences

ERL is a basic framework for evolution-guided exploration methods that has outperformed PPO, DDPG, and GA on MuJoCo continuous control benchmarks [17]. In this method, EC employs a population of agents to explore the parameter space to generate diverse experiences for the training of off-policy RL agents, and periodically copies the RL agent into the EC population to inject gradient information into evolution. As a result, ERL is able to deal with the challenges of sparse rewards, ineffective exploration, and brittle convergence properties. EC is indifferent to the reward sparsity by using an episodic fitness metric and enables diverse exploration and introduces redundant information with a population of actors. The periodic injection of gradient information of RL into EC deals with the inefficient exploration issues of EC.

A more general framework, Collaborative ERL (CERL), maintains a population of TD3 agents to optimize over different hyperparameters (e.g., discount rate $\gamma$) and applies a resource manager to allocate computational resources to the agents adaptively according to their cumulative returns [3]. CERL has outperformed ERL and TD3 in Humanoid and Swimmer tasks, where state-of-the-art RL algorithms are highly sensitive to their hyperparameters. To further improve the learning efficiency, ERL with 2-scale state representation and policy representation (ERL-Re$^2$) divides the whole policy representation into a nonlinear state representation shared by EC agents and RL agents and a linear policy representation optimized separately by all agents [194]. Moreover, the linear policy representation has enabled ERL-Re$^2$ to design behavior-level crossover and mutation operators with clear semantics. Surprisingly, ERL-Re$^2$ has achieved significant improvements over a number of ERL variants and state-of-the-art RL algorithms in MuJoCo tasks. In addition, the competitive and cooperative heterogeneous DRL (C2HRL) leverages the advantages of both gradient-based and gradient-free agents and introduces 2 agent management mechanisms to compete for computational resources and share exploration experiences [195]. C2HRL has shown faster convergence than CERL in MuJoCo tasks.

To generate legal offspring, more sophisticated variation operators such as safe mutation [133,134] and Q-filtered distillation crossovers [130] have been introduced into the proximal distilled ERL algorithm [131] and the safe-oriented search method [134]. Additionally, ERL is augmented with imitation learning, where RL agents learn from the experiences sampled by high-fitness EC individuals, and low-fitness EC individuals learn from RL agents by imitating behavior patterns [196]. This method has outperformed DDPG and ERL on 4 MuJoCo tasks.

However, the EC part of ERL applies undirected exploration by adding noise to the parameters. Hence, directed exploration methods such as NS, QD, and curiosity search have been applied to cover the state–action space more uniformly and efficiently. For example, GEP-PG applies curiosity search to generate diverse targeted samples for the training of DDPG [192]. In addition to adding noise in the parameter space, EC enables the introduction of action noise for off-policy RL algorithms. The evolutionary action selection-TD3 (EASTD3) uses samples generated by RL agents to form an EC population, applies particle swarm optimization (PSO) to evolve continuous action values, and finally uses the best actions to guide action selection for RL agents [197]. EASTD3 has shown better performance than ERL, PDERL, CERL, and TD3 on MuJoCo tasks.

Other exploration methods in robotics include evolving a foot trajectory generator to provide diversified motion priors to guide policy learning [198] and augmenting NS with multiple behavior spaces to deal with the challenge of automated data collection in robotic grasping tasks [199].

### Use of gradient information

In addition to the combinations of GAs and off-policy RL algorithms in the ERL variants, the cross-entropy method (CEM) has been combined with TD3 or DDPG to create CEM-RL [200]. In CEM-RL, the gradient information of the RL agent is directly used to half of the CEM agents at each iteration to increase training efficiency. Asynchronous ES-RL, based on CEM-RL and OpenAI ES, has been developed by Lee et al. [201] to integrate ES with off-policy RL methods and improve time efficiency and performance over ERL and CEM-RL. Since ESs are similar to gradient-based RL methods, they are naturally applied to the EC loop of ERL to share gradient information with RL methods. The combination of ES and SAC, called ESAC, enables effective exploration in the parameter space [202]. ESAC has obtained improved performance over SAC, TD3, PPO, and ES on many MuJoCo and DeepMind control suite locomotion tasks.

ERL and its variants have been applied only to the off-policy actor–critic methods. Therefore, Supe-RL has been proposed by Marchesini et al. [203] to generalize to any RL methods using soft updates for policy evolution. Supe-RL generates a set of children by adding Gaussian mutation to the policy weights, and then soft updates the best individual from the children periodically or keeps the weights the same to avoid detrimental behaviors. Supe-RL has outperformed ERL and PPO on several MuJoCo tasks. Additionally, Zhu et al. [204] have proposed the gradient-evolutionary algorithm with temporal logic for on-policy methods.

### Discussion

While QD algorithms have shown promise in encouraging diversity in neuroevolution and potentially realizing the third pillar of AI-generating algorithms (AI-GAs) [205], their effectiveness relies heavily on the selection of appropriate behavioral characteristics (BCs). Ensuring that the extraction of BCs aligns with quality is crucial to the success of QD, as otherwise, it may perform worse than other methods like NS. This highlights the importance of careful consideration and experimentation when selecting BCs for QD algorithms.

Another important consideration when using evolution-guided exploration methods in off-policy RL algorithms is the alignment between the EC population and the RL agents. If the EC population is too different from the RL agents, the experiences or gradients generated by EC may not be efficient for updating the RL agents. This underscores the need to carefully design and tune EC algorithms to ensure that they are suitable for the specific RL tasks at hand. Additionally, it may be beneficial to explore hybrid methods that combine EC and gradient-based methods to leverage the strengths of both methods.

## EC in Reward Shaping

The reward signal is crucial in reflecting the task objective in RL. However, in many scenarios, the reward is sparse, making it difficult for agents to learn useful information. To tackle this issue, reward shaping has been introduced, which enhances the original reward with additional shaping rewards. These subrewards provide feedback about the progress of the task, adjust the importance of different aspects of the task, or learn proxy rewards [206]. Empirical studies have shown that reward shaping can reduce the amount of exploration and accelerate convergence [207]. Despite these benefits, reward shaping still faces several challenges. Firstly, it can alter the problem itself. Secondly, designing appropriate subrewards requires expert domain knowledge. Thirdly, achieving a balance between multiple subrewards requires careful manual tuning. Finally, credit assignment is a difficult problem in multi-agent RL. EC has been applied to deal with these challenges by the evolution of reward functions directly and hyperparameters of parameterized rewards for both single-agent and multi-agent RL.

### Evolution of reward functions

Potential-based reward shaping, which originated from the potential energy, was proposed in the 1990s to deal with the challenge of changes in the problem itself [208]. The method derives a potential-based shaping function that can guarantee its consistency with the optimal policy. However, designing the potential function is still an open question, requiring extensive manual search to produce acceptable results. Therefore, the reward network in Ref. [209] employs potential-based reward shaping, representing the potential function as an NN whose weights are optimized by NES in a highly parallel way, as in OpenAI ES. The reward network, along with the proposed synthetic environments (both trained by NES), is robust to hyperparameter variation and can be transferred to unseen agents.

In contrast, a general computational reward framework from the evolutionary perspective achieves the optimal reward function by maximizing the expected fitness over the distribution of environments [210]. Experiments in a hungry–thirsty task have shown that the optimal reward function can capture both physical regularities across environments and specific properties of agent-environment interactions. Moreover, maximizing the expected fitness can lead to the emergence of interesting reward functions such as intrinsic and extrinsic rewards. As a result, the framework avoids the issues of designing task-dependent subrewards. However, the automated quasi-exhaustive search used for finding good reward functions is quite time-consuming. Hence, PushGP has been applied to find reward functions efficiently, which can discover common features of environments and reduce the expensive costs for sim-to-real transfer in robot control tasks [211]. Other EC methods to obtain intrinsic rewards include hyperparameter tuning for reward shaping [212], symbolic reward search [213], and exploration methods such as NS [214] and curiosity search [163] (demonstrated in the EC in Exploration section).

In a multi-agent Cyber Rodent robots task, the parameters of an intrinsic reward function are evolved to dynamically control the exploration for multiple hand-coded extrinsic rewards [212]. In several MuJoCo and Atari tasks, dense intrinsic reward functions are evolved directly by symbolic regression for the purpose of obtaining interpretable low-dimensional reward functions [213]. In a grounded communication environment, the multi-agent evolutionary reinforcement learning (MERL) method maintains a population of evolving teams by neuroevolutionary algorithms to achieve sparse team-based rewards and optimizes agent-specific policies by gradient-based methods to achieve a local reward for each agent [215]. MERL has outperformed the state-of-the-art MARL algorithm MADDPG [216]. In tasks with both spatial and temporal constraints (i.e., a team of agents should complete a task in temporal and spatial order), the multi-agent evolution via dynamic skill selection (MAEDyS) method first decomposes a task into subcomponents with local rewards and then applies a coevolutionary method to optimize multiple local rewards and a global reward [217].

## Parameterization

Generally, dense rewards can be introduced by parameterizing rewards or balancing the relative importance between internal rewards. However, the traditional manual tuning methods to design these dense rewards require too much prior knowledge and are low efficient. These issues can be addressed through HPO methods. For example, the AutoRL method applies large-scale HPO to the parameterized reward function and the NN architecture of the policy [218]. The method first learns the reward function using evolutionary strategies with fixed NN architectures of the actor and critic, and then learns NN architectures by only adjusting the number of neurons in each layer with the previously learned reward. Although it can find robust policies for point-to-point robot navigation tasks, it is still relatively inefficient. Besides, AutoRL has been applied to several MuJoCo tasks and outperformed state-of-the-art algorithms such as SAC and PPO [206]. In addition, in Ref. [57], a number of weights in the hand-designed potential-based reward shaping is evolved together with other hyperparameters to control the trade-off between exploration and exploitation in a robot foraging task.

The credit assignment issue in tasks with sparse rewards is quite challenging, as estimating the contribution of an agent is particularly hard, making it difficult to optimize team rewards. A number of works have dealt with this issue using HPO methods. The state-of-the-art PBT has been used to automatically learn dense internal rewards for the popular 3D multiplayer first-person video game [58] and optimize the relative importance between a set of dense shaping rewards along with their discount rates automatically for the continuous multi-agent MuJoCo soccer game [60]. Both methods aim to align the myopic shaping rewards with the sparse long-horizon team rewards and generate cooperative behaviors. In addition, Wang et al. [219] deal with intertemporal social dilemmas by trading off collective welfare and individual utility. Specifically, a shared intrinsic reward network takes features input from all agents, while each agent trains a distinct policy network in each episode. Then, PBT is applied to optimize the weights of the reward network and other hyperparameters to evolve altruistic behavior.

## Discussion

The use of EC for reward shaping has shown promising results in addressing the challenges of sparse rewards in RL. Potential-based reward shaping, a method originating from potential energy, can guarantee consistency with the optimal policy. A computational framework can achieve optimal reward functions through a quasi-exhaustive search, and PushGP can efficiently find reward functions that discover common features of environments. HPO methods, such as AutoRL, can be used for parameterized reward functions and NN architecture tuning. In multi-agent RL, PBT has been used to automatically learn dense internal rewards and optimize the relative importance between shaping rewards to generate cooperative behaviors.

Although EC methods offer an automated and efficient approach to reward shaping in RL, relying solely on EC methods for reward shaping can be inefficient. To mitigate this, incorporating prior knowledge can be beneficial. Intrinsic rewards can guide agents to explore interesting parts of the state space, while extrinsic rewards define the task's objective. EC methods can generate intrinsic rewards and dynamically tune their weights to address the issue of sparse rewards.

## EC in Meta-RL

RL has proven successful in tackling complex tasks, but it often requires a large number of samples to learn from scratch for each task. Moreover, the choice of a pre-specified RL algorithm can impact the performance, such as cumulative rewards or sample efficiency. To address these challenges, meta-RL seeks to develop a general-purpose learning algorithm that can adapt to different tasks. In other words, it aims to leverage knowledge from previous tasks to facilitate fast learning in new ones. Meta-RL can handle various scenarios, including learning in similar environments from a single task or substantially distinct environments from multiple tasks.

From the perspective of optimization-based methods, meta-RL can be formulated as a bi-level optimization problem, where the inner level involves learning an agent using standard RL techniques, while the outer level optimizes RL configurations such as policy update rules, hyperparameters, and reward formulation to achieve a meta-objective. Gradient-based methods [220], RL methods [221], and EC methods [222] can be used to optimize either level. EC methods are particularly promising for meta-RL since they are applicable for non-differentiable meta-objectives and can avoid the high computational overhead of high-order gradients. Specifically, EC methods have been introduced in various aspects of meta-RL, such as parameter initialization, loss learning, environment synthesis, and algorithm generation.

## Parameter initialization

Parameter initialization is a critical aspect of meta-RL that aims to find a suitable policy initialization such that good general performance can be achieved with only a few gradient steps over different environments sampled from a distribution of tasks. The model-agnostic meta-learning (MAML) is a state-of-the-art method that formulates the learning of an easily adaptable policy as an optimization problem with a meta-objective of minimizing the loss of a small number of gradient steps on a new task [220]. MAML has no requirement for the model representation and has been applied successfully to various tasks, including regression, classification, and RL. However, estimating the second derivatives of the reward function is challenging using back-propagation, and policy gradient methods have inherently high variance. To overcome these limitations, ES-MAML integrates MAML into the ES framework, which avoids calculating the second derivatives by Gaussian smoothing of the MAML reward [223]. Additionally, ES-MAML simultaneously optimizes hyperparameters and initial parameters, leading to better performance and exploration on tasks with sparse rewards. Alternatively, the Baldwin evolutionary methods, which involve reinitializing parameters and hyperparameters, have been used for meta-learning when MAML is not applicable [224]. However, it is worth noting that meta-learning parameter initialization is limited to a single task or similar distribution of tasks.

## Loss learning

Meta-learning the loss has demonstrated generalization ability across substantially different tasks, including out-of-distribution tasks. Evolved policy gradient (EPG) is a representative work that meta-learns a differentiable loss function parameterized by temporal experiences [222]. EPG has 2 optimization loops: an inner loop learns a policy to minimize the loss, and an outer loop learns the loss such that an agent trained by the loss can achieve high expected returns in a task distribution. The parameters of the loss function, represented by an NN, are optimized by OpenAI ES, where a population of workers runs in parallel to obtain the update gradients of the loss. EPG has demonstrated better generalization than MAML on several MuJoCo tasks, although its effectiveness is limited to a small family of tasks at a time.

Several works have meta-learned interpretable loss functions by symbolic regression of GP. For example, Co-Reyes et al. [225] generate symbolic loss functions for general policy update rules by regularized evolution and directly evolve a population of RL algorithms. Regularized evolution removes the oldest solutions to prevent overfitting to training noise, rather than removing the worst solutions from the whole population, such that algorithms retrained well are more likely to remain in the population. Experiments on complex tasks demonstrate that, when learning from scratch, this method can rediscover DQN, and when inserting the existing algorithm DQN into the population as bootstrap, the method can further improve generalization of DQN.

Additionally, to simultaneously meet the requirements of performance, generalizability, and stability, MetaPG formulates the requirements as 3 meta-objectives and applies NSGA-II to discover new RL algorithms by designing directed acyclic graphs, a symbolic representation [226]. Experiments on 3 continuous control tasks show that MetaPG has improved both the performance and generalizability of SAC by using a graph-based implementation of SAC to initialize the population [32].

### Environment synthesis

Environment synthesis aims to generate synthetic data that can enhance the training efficiency of RL models. In the world models approach, for instance, the MDN-RNN model receives compressed input from the agent's visual perception and predicts future states to improve training efficiency [98]. However, in real-world applications, it is essential to consider both states and rewards. Therefore, Ferreira et al. [209] propose a method that simultaneously learns synthetic environments (SEs) and reward networks (RNs) to generate synthetic MDPs that mimic real-world environments. The method represents SEs and RNs as NN proxies and applies a bi-level optimization method to optimize them for better performance on real environments. The inner optimization trains RL agents on the proxies, while the outer loop evolves parameters of SEs and RNs using NES to maximize performance on real environments. This method allows training competitive agents with fewer interactions with real environments on MuJoCo tasks. In Ref. [58], the PBT [51] has been applied to meta-learn the internal rewards and hyperparameters of the RL algorithm simultaneously. This method can be viewed as a 2-tier optimization problem where the inner tier maximizes the expected discounted internal rewards, and the outer tier maximizes a meta-reward based on the internal rewards and hyperparameters. The joint optimization of policy and the RL process itself enables the training of better agents in large-scale complex tasks.

### Algorithm generation

When rewards from extrinsic environments are sparse, curiosity can provide intrinsic motivation for agents to explore environments. In Ref. [227], learning how to explore is formulated as a meta-learning problem of generating curious behaviors, where effective curiosity algorithms can be generated by learning proxy rewards for exploration. The meta-learning method has 2 optimization loops: the outer loop evolves a population of curiosity algorithms (represented as programs) by GP to learn intrinsic reward signals dynamically, and the inner loop performs a standard RL pipeline using the learned reward signals. The method designs a domain-specific language, represented as directed acyclic graphs, to generate programs that include building blocks such as NNs, objective functions, ensembles, buffers, and other regressors as polymorphic data types. Empirical studies have shown that the method can discover algorithms similar to NS and generalize across a much broader distribution of environments.

### Discussion

Meta-RL has undoubtedly made significant progress in enabling RL to learn new tasks efficiently and generalize across different tasks. However, the number and complexity of tasks that can be solved using meta-RL are still limited, particularly in real-world applications. Furthermore, the computational cost of meta-RL is high due to the concurrent optimization of 2 loops and training over a large number of tasks. Therefore, exploiting the model-agnostic and highly parallel properties of EC is a promising direction to unlock the full potential of meta-RL. EC methods can avoid the high computational overhead of high-order gradients and can handle non-differentiable meta-objectives, which are challenging for gradient-based methods. The use of EC methods in various aspects of meta-RL, such as parameter initialization, loss learning, environment synthesis, and algorithm generation, has demonstrated promising results. With further research and development, EC-based meta-RL has the potential to enable RL to learn and generalize across more complex tasks and to be more computationally efficient in real-world scenarios.

Although EC methods have shown great potential in meta-RL, there are still several challenges to overcome. One challenge is the high dimensionality of the search space, which can be very large, especially when dealing with complex environments or when optimizing both policy and hyperparameters. This can lead to slow convergence and high computational costs. Another challenge is finding a balance between exploration and exploitation. EC methods often rely on some form of exploration to find good solutions, but too much exploration can result in wasted computational resources and slow progress. To address these challenges, future research in EC-based meta-RL could focus on developing more efficient search algorithms, reducing the dimensionality of the search space, and designing exploration strategies that balance exploration and exploitation more effectively.

## EC in Multi-Objective RL

While the aforementioned methods focus on single-objective RL, many real-world problems have multiple conflicting objectives. For example, an agent may need to grasp an object while minimizing its energy consumption. These problems can be formulated as multi-objective MDPs (MOMDPs), where the reward function $R = [r_1, ..., r_m]^T$ represents a vector of $m$ rewards, and the discount factor is a vector $\gamma = [\gamma_1, ..., \gamma_m]^T$. Multi-objective RL (MORL) learns a policy $\pi_\theta$ to simultaneously optimize the multiple objectives $J(\pi) = [J_1(\pi), ..., J_m(\pi)]^T$, where each objective $J_i(\pi)$ is associated with a dimension of the reward vector:

$$J_i(\pi) = \mathbb{E}_{\rho_0, \pi, T}\left[\sum_{t=0}^{\infty} \gamma_i^t r_i^t\right]. \qquad (2)$$

Therefore, MORL is intrinsically a multi-objective optimization problem, where there is no single optimal solution that can satisfy multiple conflicting objectives. Instead, a set of trade-off solutions named Pareto optimality is desired. In this context, Pareto optimality defines a solution that is not dominated by any other solution in the objective space. The set of such solutions is called the Pareto set, and the mapping of the Pareto set in the objective space is referred to as the Pareto front. To find a set of solutions approximating the Pareto front, multi-objective evolutionary algorithms (MOEAs) are effective tools to obtain optimal solutions in a single run [228]. The solution qualities are often justified based on 2 aspects: convergence toward the true Pareto front and diversity along the Pareto front [229]. The hypervolume (HV) metric is often employed to measure convergence and diversity concurrently, which calculates the size of the hypervolume enclosed by the solution set and a reference point (e.g., a vector of nadir points of all obtained solutions in each dimension) [230]. Thus, a larger HV metric value indicates a closer approximation of the Pareto front of solutions. Other metrics employed in MORL include the (inverted) generational distance [231], the generalized spread indicator [232], the cardinality indicator [233], and sparsity metrics [234].

According to the number of policies obtained at the end of optimization, MORL methods can be roughly divided into 2 categories: single-policy methods and multi-policy methods [23]. Single-policy methods aim to learn a unique optimal policy each time. To achieve this, multiple rewards are transformed into a scalar reward by using a scalarization function, and then the scalar-rewarded task is learned by general RL methods [235]. However, single-policy methods have several drawbacks, including the requirement for domain knowledge, low efficiency, and sub-optimality due to preference [23]. In contrast, multi-policy methods aim to find a set of diverse policies that approximate the Pareto front. These methods learn a set of policies that provide users with multiple options to choose from. One class of multi-policy methods focuses on how to apply the weights of objectives to guide the optimization of policies [236]. However, scalarizing for all weights and approximating the whole Pareto front are both challenging. In contrast, another class of multi-policy methods applies MOEAs to obtain a set of optimal solutions without setting weights. Further, multi-objectivization is used to transfer single-objective problems into multi-objective problems to make problems easier to solve.

## Multi-objective evolutionary algorithms
To deal with the existing issues of MORL, EC methods are effective tools to obtain a set of trade-off solutions that can not only converge toward the Pareto front but also distribute well. In Ref. [19], NSGA-II or MCMA (a multi-objective variant of CMA-ES) augmented with a local search method has been applied to search the graph-represented policies in robot load tasks with 2, 3, and 5 objectives. In addition, evaluation metrics have been applied to select solutions with good performance to enter into the next generation. The metrics-based MORL method applies HV and sparsity metrics to select the best policy–weight pairs [234]. With a prediction model to predict policy improvement and an intra-family interpolation method to construct continuous Pareto fronts, the method has achieved better HV and sparsity metrics over MOEA/D and a meta-learning method on a set of multi-objective MuJoCo tasks. In addition, the ideas of selecting good performance by metrics have been borrowed into the action selection mechanism in MORL. The hypervolume-based MORL algorithm selects the action with the largest HV contribution into the next generation, and this method has outperformed a linear scalarization-based action selection mechanism on 2-objective deep sea treasure and 3-objective mountain car benchmark tasks [237]. Besides, HV has been applied to select actions in an interactive way [238]. The Pareto-Q-learning algorithm used 3 evaluation methods, i.e., Pareto dominance relation, HV metric, and the cardinality metric, to select the most potential actions in the Q-learning process [239].

## Multi-objectivization
Multi-objectivization is a technique that transforms a single-objective problem into a multi-objective problem by decomposing a single objective or adding extra objectives [240]. The motivation behind this approach is to make the problem easier to solve by introducing more exploration through prior knowledge. In the context of RL, multi-objectivization is typically used as a reward shaping method that focuses on not only completing tasks but also encouraging behavior diversity among learned agents. Recently, the evolutionary multi-objective game AI (EMOGI) framework was proposed as a method for generating behavior-diverse game agents [241]. EMOGI first tailors a reward function with multiple objectives consisting of a performance objective and a number of behavior objectives, and then applies NSGA-II to evolve a set of trade-off policies. Further studies on EMOGI have shown that introducing behavior objectives alone, without the performance objective, is able to generate a set of behavior-diverse agents [242].

## Discussion
EC has a wide range of MOEAs such as Pareto-based, decomposition-based, and indicator-based methods, but MORL has mainly used NSGA-II and the HV metric. The lack of diversity in MOEAs for MORL has limited the development of efficient and effective algorithms for multi-objective RL problems. One possible solution to address this issue is to introduce user preference-based MORL methods. Such methods can take inspiration from preference-based MOEAs and enable users to specify their preferences for different objectives. Furthermore, to advance the field of MORL, researchers need to design and study tasks with more than 3 objectives. This would allow the use of many-objective evolutionary algorithms (MaOEAs) [243], which have the potential to provide better solutions for such problems.

While EC has proved to be a valuable tool for MORL, there is still a need to explore more MOEAs and essential ideas to develop efficient and effective algorithms for multi-objective RL problems. Preference-based MORL methods and MaOEAs can offer new opportunities to overcome the limitations of existing approaches and advance the field further. Future research efforts should focus on designing and studying more complex tasks with multiple objectives to better evaluate the performance of MORL algorithms and compare them with existing state-of-the-art methods.

## Future Research Directions

Although EvoRL has been successfully applied to large-scale complex RL tasks, even with sparse and deceptive rewards, it is still computationally expensive. A number of efficient methods in terms of EvoRL processes including encoding, sampling, search operators, algorithmic framework, and evaluation, as well as benchmarks, platforms, and applications, are desirable research directions, as overviewed in Fig. 4.

### Encodings

Efficient encodings of decision variables are essential for the optimization of EC algorithms. While real encoding is the most widely used representation in most research fields of RL, it is not the most efficient representation for policy search. Indirect encodings, such as CPPN, can be easily evolved to capture structural relationships that exist in the human body structure. Additionally, TPG for GP is a highly compressed representation that organizes multiple execution programs into modular structures. Despite these indirect encodings, the development of efficient encoding schemes is still limited. This includes the development of human-understandable encodings and efficient encodings applied to ESs. Furthermore, it is desirable to apply existing encodings to large-scale complex tasks to investigate their scalability and generality.

Efficient encodings have the potential to improve the scalability and generality of EC algorithms. It is crucial to develop efficient encoding schemes that can handle large-scale complex tasks to reduce computational costs. Furthermore, there is a need for human-understandable encodings that can increase the interpretability of evolved policies. The development of efficient encoding schemes can also benefit from incorporating insights from other research fields, such as information theory and statistical learning theory. Efficient encoding schemes can also be combined with other efficient methods, such as efficient sampling and efficient sample utilization, to further improve the performance of EC algorithms. Overall, the development of efficient representations is an important research direction for improving the scalability and generality of EC algorithms.

### Sampling methods

The fundamental process of EC is to continuously sample solutions in search of the optimal direction. Therefore, efficient sampling can significantly reduce time and computational overhead. In ES-based methods for policy search, efficient sampling can decrease variance and accelerate convergence, such as mirrored sampling in OpenAI ES [16]. Efficient sampling from the decision space is advantageous for ESs, but it has not been thoroughly investigated in GA- or GP-based methods.

On the other hand, in GA-based methods, sampling is often conducted based on the performance in the objective space (i.e., behavior space). For instance, in exploration, new individuals are frequently generated in sparsely populated areas to help find agents with diverse behaviors as quickly as possible. However, designing efficient sampling methods remains a challenging issue for scaling up algorithms to tasks with large-scale search spaces, sparse and deceptive rewards, and complex and varied landscapes during training.

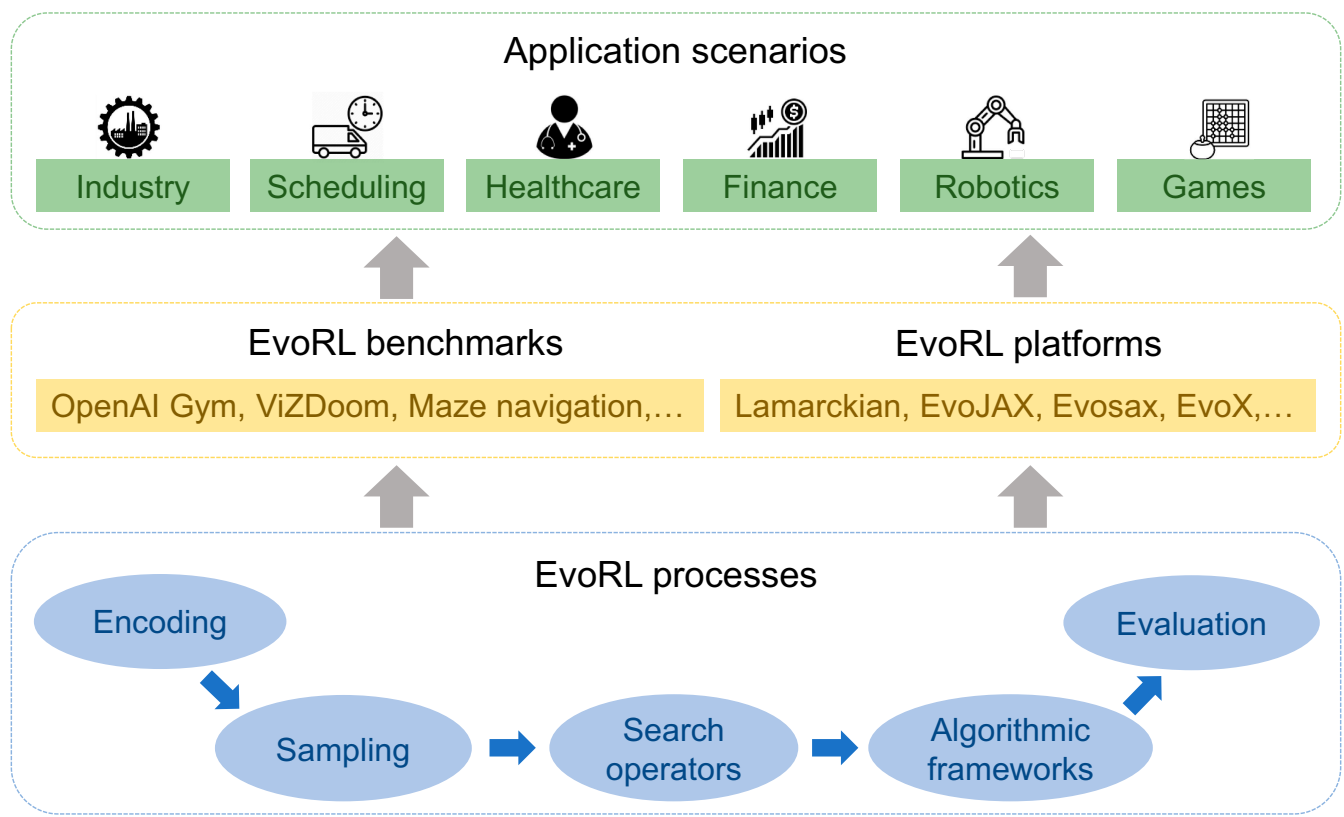

**Fig. 4.** Overview of future directions from 4 fields of EvoRL: processes, benchmarks, platforms, and application scenarios.

## Sample utilization
In off-policy RL algorithms such as DQN, samples are stored in a replay buffer for multiple epochs of updates to improve sample utilization and break sample correlations. On-policy RL algorithms such as PPO use importance sampling to enable sample reuse by correcting for the discrepancy between old and new policies. Other RL techniques such as generalized advantage estimation and V-trace allow the reuse of old samples to further reduce variance in updates [244].

In ESs (similar to policy-gradient RL methods), techniques such as importance sampling have been introduced to enable sample reuse [76]. However, importance sampling alone is not sufficient to compete with RL methods in terms of sample efficiency. Hence, more efficient techniques need to be introduced into ESs. Furthermore, it remains to be investigated whether these techniques are feasible in GAs.

## Search operators
Although there have been a number of variational operators tailored for the search of NNs, such as safe mutation [133] and distillation crossover [131], the related research is still limited due to the intrinsic features of various encodings. Besides topology distillation and gradient-sensitive variation through imitation learning and distillation, more methods or techniques can be employed to improve the efficiency of variational operators, such as transfer learning and surrogate gradient methods.

In addition, the operators need to be customized according to the encoding method and can be controlled to generate safe offspring. To do this, the use of surrogate models or other auxiliary tools can help identify the most promising offspring to preserve during the selection process, while discarding less promising ones. Finally, more search operators can be borrowed from EC (e.g., DE [245], PSO [246], and CSO [247]) to investigate their effectiveness for RL tasks. The development of efficient search operators can significantly reduce the search time and improve the effectiveness of EvoRL algorithms.

## Algorithmic frameworks
Several EvoRL frameworks have been developed, such as NEAT, OpenAI ES, ERL, and PBT. However, there are still several issues that require further research. In policy search, optimizing a policy with a large number of parameters can be challenging. If the policy can be divided into components at a finer granularity, various functional modules can be automatically discovered using cooperative cooperation methods. Moreover, this divide-and-conquer method enables parallel computation of different components, thereby further reducing computational resources.

Research on evolution-guided exploration methods has mainly focused on off-policy RL algorithms since experiences of EC are more easily used than gradient information to ensure RL algorithms do not degenerate. If gradient information of EC can be introduced, the training efficiency of RL algorithms can be improved. Additionally, more effort is needed to develop frameworks for on-policy RL algorithms.

Although PBT is effective, selecting optimized hyperparameters from a large number of hyperparameters is challenging. This issue can be formulated as a combinatorial optimization problem, and then EC can be applied to the whole framework of hyperparameter selection and optimization to realize end-to-end automated HPO.

## Evaluation methods
Efficient evaluation methods are crucial to reduce the computational burden of evaluating the fitness of each newly generated agent in EvoRL. Surrogate-assisted methods have been introduced in EvoRL to predict fitness values [248,249]. However, accurately modeling the relationship between policy parameters/behaviors and performance remains a challenging issue, especially in tasks with a large number of parameters and multiple performance indicators. In addition, ranking agents without fitness estimation is a feasible method that has been explored, such as rank-based fitness shaping that introduced racing methods into ESs to judge the relative performance of agents [83]. These existing attempts have shown that accurate fitness values are not always necessary for efficient evaluation. A partial order relation among agents may be sufficient, further facilitating the parallelism of algorithms.

Efficient evaluation methods are essential not only for speeding up the evaluation process but also for improving the accuracy and robustness of evaluation results. In addition to surrogate-assisted methods and rank-based fitness shaping, other techniques such as transfer learning, meta-learning, and Bayesian optimization can be explored to reduce the number of interactions with the environment and improve the efficiency of evaluation. These techniques can also be used to improve the generalization capability of the agents, making them more adaptable to new tasks and environments. Efficient evaluation methods can significantly improve the scalability and generality of EvoRL algorithms, enabling them to tackle more complex and challenging tasks in a more efficient manner.

## Benchmarks for EvoRL
The earlier tasks resolved by EvoRL are simple, such as pole balancing tasks and mountain car tasks [82]. Then, EvoRL has been applied to various mobile robotics tasks such as maze navigation and robot arm control, as well as sim-to-real robotic tasks. These works have contributed to the formation of the research field of evolutionary robotics [166]. Recently, benefiting from the unified platform and agent–environment interaction interface of integrated OpenAI Gym [250], games have become excellent benchmarks for evaluating EvoRL compared to existing testbeds. In particular, EvoRL has been widely applied to tasks with continuous state–action space (e.g., robot control in MuJoCo), and has shown promise for learning agents in large-scale visual-input games (e.g., ViZDoom [150] and Dota2 [151]).

However, in empirical studies, EvoRL methods have typically been compared with RL methods to observe their performance improvement, rather than compared with other EvoRL methods. The reason is that existing RL benchmarks are not sufficient to investigate the properties of various EvoRL methods. On the other hand, multi-objective EvoRL methods and QD EvoRL methods do not have tailored benchmarks. Therefore, developing tailored EvoRL benchmarks is much needed. While designing EvoRL benchmarks is not easy, a quick way is to modify existing RL benchmarks, such as the modified multi-objective

MuJoCo tasks [7] to verify multi-objective EvoRL methods and the 2D bipedal walking adapted tasks to verify the open-endedness of EvoRL methods [179].

## Scalable platforms

Recently, several scalable platforms for EvoRL have been developed, such as the Lamarckian platform, an open-source high-performance platform that can scale up to thousands of CPU cores and has been verified to perform well in large-scale commercial games [251]. Another platform is the parallel evolutionary and reinforcement learning library (PEARL), although it does not show highly scalable abilities [252]. With the fast computational ability of GPUs, the JAX library has been released, which offers a NumPy-like API for GPU-accelerated numerical calculations. Based on JAX, the platforms for neuroevolution such as EvoJAX [253] and evosax [254], the platform EvoX [255] for general EC algorithms, and the platform QDax [256] for QD algorithms have been developed. These platforms have shown the ability to find solutions in Atari or MuJoCo tasks with significantly less time than using CPUs.

In addition, Bhatia et al. [257] have proposed a platform for developing co-evolution algorithms for co-optimizing the design and control of robots. Despite these advancements, the research on efficient and scalable platforms for EvoRL is still limited, as they may not be user-friendly or focus on limited EvoRL algorithms. Further research is needed to develop more efficient and scalable platforms that can handle large-scale complex tasks and integrate with various EvoRL algorithms.

## Conclusion

This article presents a comprehensive survey of EvoRL, primarily focusing on its methodologies and future directions. Firstly, the article introduces EvoRL methods by categorizing them into 6 key research fields of RL: HPO, policy search, exploration, reward shaping, meta-RL, and multi-objective RL. For each field, the applied EC methods, including ESs, GAs, and GP, are explained, and their main advantages and disadvantages are discussed. Secondly, the article explores several future directions for developing efficient methods in EvoRL processes, as well as tailored EvoRL benchmarks and platforms. By discussing these future directions, the article offers guidance for researchers and practitioners interested in the field of EvoRL, and fosters the continued growth of this cross-disciplinary research area. In conclusion, this survey serves as a valuable resource for anyone interested in learning about EvoRL and its potential applications in RL, and provides insights into the future trajectory of this rapidly expanding field.

## Acknowledgments

**Funding:** This work is supported by the Program for Guangdong Introducing Innovative and Entrepreneurial Teams (Grant No. 2017ZT07X386). Y.J. is led by an Alexander von Humboldt Professor for Artificial Intelligence endowed by the German Federal Ministry of Education and Research. **Author contributions:** Both H.B. and R.C. wrote the manuscript. Y.J. contributed to the final version of the manuscript. R.C. supervised the project. **Competing interests:** The authors declare that they have no competing interests.

## Data Availability

No new data were created or analyzed in this study. Data sharing is not applicable to this article.

## References

1. Sutton RS, Barto AG. *Reinforcement learning: An introduction*; Cambridge (MA)/London (England): MIT Press; 2018.

2. Mnih V, Kavukcuoglu K, Silver D, Rusu AA, Veness J, Bellemare MG, Graves A, Riedmiller M, Fidjeland AK, Ostrovski G, et al. Human-level control through deep reinforcement learning. *Nature*. 2015;518(7540):529–533.

3. Khadka S, Majumdar S, Nassar T, Dwiel Z, Tumer E, Miret S, Liu Y, Tumer K. Collaborative evolutionary reinforcement learning. Paper presented at: Proceedings of the 36th International Conference on Machine Learning; 2019 May 24; Long Beach, CA.

4. Ecoffet A, Huizinga J, Lehman J, Stanley KO, Clune J. Go-explore: A new approach for hard-exploration problems. arXiv. 2019. https://doi.org/10.48550/arXiv.1901.10995

5. Long Q, Zhou Z, Gupta A, Fang F, Wu Y, Wang X. Evolutionary population curriculum for scaling multi-agent reinforcement learning. Paper presented at: International Conference on Learning Representations; 2020 Apr 26; Virtual conference.

6. Conti E, Madhavan V, Petroski Such F, Lehman J, Stanley K, Clune J. Improving exploration in evolution strategies for deep reinforcement learning via a population of novelty-seeking agents. *Adv Neural Inf Proces Syst*. 2018;31.

7. Roijers DM, Vamplew P, Whiteson S, Dazeley R. A survey of multi-objective sequential decision-making. *J Artif Intell Res*. 2013;48:67–113.

8. Ebrahimi S, Rohrbach A, Darrell T. Gradient-free policy architecture search and adaptation. Paper presented at: Proceedings of the 1st Conference on Robot Learning (CoRL); 2017 Oct 18; Mountain View, CA.

9. Wang Z, Chen C, Dong D. Instance weighted incremental evolution strategies for reinforcement learning in dynamic environments. *IEEE Trans Neural Netw Learn Syst*. 2022;1–15.

10. Luo Z-Q, Yu W. An introduction to convex optimization for communications and signal processing. *IEEE J Sel Areas Commun*. 2006;24(8):1426–1438.

11. Pereyra M, Schniter P, Chouzenoux E, Pesquet J-C, Tourneret J-Y, Hero AO, McLaughlin S. A survey of stochastic simulation and optimization methods in signal processing. *IEEE J Sel Top Signal Process*. 2016;10(2):224–241.

12. Tian Y, Si L, Zhang X, Cheng R, He C, Tan K, Jin Y. Evolutionary large-scale multi-objective optimization: A survey. *ACM Comput Surv*. 2021;54(8):1–34.

13. Yazdani D, Cheng R, Yazdani D, Branke J, Jin Y, Yao X. A survey of evolutionary continuous dynamic optimization over two decades—Part B. *IEEE Trans Evol Comput*. 2021;25(4):630–650.

14. Lin X, Yang Z, Zhang Q. Pareto set learning for neural multi-objective combinatorial optimization. Paper presented at: International Conference on Learning Representations; 2022 Apr 25, Virtual conference; https://openreview.net/forum?id=QuObT9BTWo.

15. Li L, He C, Xu W, Pan L. Pioneer selection for evolutionary multiobjective optimization with discontinuous feasible region. *Swarm Evol Comput*. 2021;65:100932.

16. Salimans T, Ho J, Chen X, Sidor S, Sutskever I. Evolution strategies as a scalable alternative to reinforcement learning. arXiv. 2017.https://doi.org/10.48550/arXiv.1703.03864

17. Khadka S, Tumer K. Evolution-guided policy gradient in reinforcement learning. Paper presented at: Proceedings of the 32nd Conference on Neural Information Processing Systems (NeurIPS 2018); 2018 Dec 2; Montréal, Canada.

18. Cully A, Clune J, Tarapore D, Mouret J-B. Robots that can adapt like animals. *Nature*. 2015;521(7553):503.

19. Soh H, Demiris Y. Evolving policies for multi-reward partially observable markov decision processes (MR-POMDPs). Paper presented at: Proceedings of the 13th Annual Conference on Genetic and Evolutionary Computation; 2011 July; Dublin, Ireland.

20. Whitley D, Dominic S, Das R, Anderson CW. Genetic reinforcement learning for neurocontrol problems. *Mach Learn*. 1993;13(2):259–284.

21. Stanley KO, Miikkulainen R. Evolving neural networks through augmenting topologies. *Evol Comput*. 2002;10(2):99–127.

22. Sigaud O. Combining evolution and deep reinforcement learning for policy search: A survey. arXiv. 2022. https://doi.org/10.48550/arXiv.2203.14009

23. Liu C, Xu X, Hu D. Multiobjective reinforcement learning: A comprehensive overview. *IEEE Trans Syst Man Cybern Syst*. 2014;45(3):385–398.

24. J. Parker-Holder, R. Rajan, X. Song, A. Biedenkapp, Y. Miao, T. Eimer, B. Zhang, V. Nguyen, R. Calandra, A. Faust, *et al*. Automated reinforcement learning (autorl): A survey and open problems. arXiv. 2022. https://doi.org/10.48550/arXiv.2201.03916

25. Qian H, Yu Y. Derivative-free reinforcement learning: A review. *Front Comp Sci*. 2021.

26. Li Y. Deep reinforcement learning: An overview. arXiv. 2018. https://doi.org/10.48550/arXiv.1701.07274

27. Schulman J, Levine S, Abbeel P, Jordan M, Moritz P. Trust region policy optimization. Paper presented at: International Conference on Machine Learning. PMLR; 2015 Jul 6; Lille, France.

28. Schulman J, Wolski F, Dhariwal P, Radford A, Klimov O. Proximal policy optimization algorithms. arXiv. 2017. https://doi.org/10.48550/arXiv.1707.06347

29. Mnih V, Badia AP, Mirza M, Graves A, Lillicrap T, Harley T, Silver D, Kavukcuoglu K. Asynchronous methods for deep reinforcement learning. Paper presented at: Proceedings of the 33rd International Conference on Machine Learning; 2016 Jun 19; New York, NY.

30. Lillicrap TP, Hunt JJ, Pritzel A, Heess N, Erez T, Tassa Y, Silver D, Wierstra D. Continuous control with deep reinforcement learning. Paper presented at: International Conference on Learning Representations; 2016 May 2–4; Caribe Hilton, San Juan, Puerto Rico.

31. Fujimoto S, Hoof H, Meger D. Addressing function approximation error in actor-critic methods. Paper presented at: Proceedings of the 35th International Conference on Machine Learning; 2018 Jul 10–15; Stockholm, Sweden.

32. Haarnoja T, Zhou A, Abbeel P, Levine S. Soft actor-critic: Off-policy maximum entropy deep reinforcement learning with a stochastic actor. Paper presented at: International Conference on Machine Learning. PMLR; 2018 Jul 10–15.

33. Mnih V, Kavukcuoglu K, Silver D, Graves A, Antonoglou I, Wierstra D, Riedmiller M. Playing atari with deep reinforcement learning. arXiv 2013. https://doi.org/10.48550/arXiv.1312.5602

34. Hessel M, Modayil J, Van Hasselt H, Schaul T, Ostrovski G, Dabney W, Horgan D, Piot B, Azar M, Silver D, Rainbow: Combining improvements in deep reinforcement learning. Paper presented at: Proceedings of the Thirty-Second AAAI Conference on Artificial Intelligence and Thirtieth Innovative Applications of Artificial Intelligence Conference and Eighth AAAI Symposium on Educational Advances in Artificial; 2018 Feb; New Orleans, LA.

35. Van Hasselt H, Guez A, Silver D. Deep reinforcement learning with double Q-learning. Paper presented at: Proceedings of the Thirtieth AAAI Conference on Artificial Intelligence; 2016 Feb; Phoenix, AZ.

36. Hansen N, Arnold DV, Auger A. Evolution strategies. In: *Springer handbook of computational intelligence*. Verlag Berlin Heidelberg: Springer; 2015, pp. 871–898.

37. Whitley D. A genetic algorithm tutorial. *Stat Comput*. 1994;4(2):65–85.

38. Burke EK, Gustafson S, Kendall G. Diversity in genetic programming: An analysis of measures and correlation with fitness. *IEEE Trans Evol Comput*. 2004;8(1):47–62.

39. Rudolph G. *Convergence properties of evolutionary algorithms*; Verlag Dr. Kovač; 1997.

40. Hansen N, Müller SD, Koumoutsakos P. Reducing the time complexity of the derandomized evolution strategy with covariance matrix adaptation (CMA-ES). *Evol Comput*. 2003;11(1):1–18.

41. Wierstra D, Schaul T, Glasmachers T, Sun Y, Peters J, Schmidhuber J. Natural evolution strategies. *J Mach Learn Res*. 2014;15(1):949–980.

42. Amari S, Douglas SC. Why natural gradient?" Paper presented at: IEEE: Proceedings of the 1998 IEEE International Conference on Acoustics, Speech and Signal Processing, ICASSP '98 (Cat. No.98CH36181); 1998 May 15; Seattle, WA.

43. Gauci J, Stanley KO. Indirect encoding of neural networks for scalable go. Paper presented at: International Conference on Parallel Problem Solving from Nature; 2010 Sep 11–15; Krakow, Poland.

44. Risi S, Togelius J. Neuroevolution in games: State of the art and open challenges. *IEEE Trans Comput Intell AI Games*. 2015;(99):1.

45. Buk Z, Koutník J, Šnorek M. Neat in hyperneat substituted with genetic programming. Paper presented at: International Conference on Adaptive and Natural Computing Algorithms; 2009 Apr 23–25; Kuopio, Finland.

46. Moraglio A, Di Chio C, Togelius J, Poli R. Geometric particle swarm optimization. *J Artif Evol Appl*. 2008;2008:143624.

47. McKay RI, Hoai NX, Whigham PA, Shan Y. Grammar-based genetic programming: A survey. *Genet Program Evolvable Mach*. 2010;11(3):365–396.

48. Deb K. *Multi-objective optimization using evolutionary algorithms*; ed. 1; Wiley-Interscience series in systems and optimization; Chichester (NY): John Wiley & Sons; 2001.

49. Lehman J, Stanley KO. Abandoning objectives: Evolution through the search for novelty alone. *Evol Comput*. 2011;19(2):189–223.

50. Zhao W, Queralta JP, Westerlund T. Sim-to-real transfer in deep reinforcement learning for robotics: A survey. *IEEE Symp Ser Comput Intell*. 2020;2020:737–744.

51. Jaderberg M, Dalibard V, Osindero S, Czarnecki WM, Donahue J, Razavi A, Vinyals O, Green T, Dunning I, Simonyan K, et al. Population based training of neural networks. arXiv. 2017. https://doi.org/10.48550/arXiv.1711.09846

52. Franke JK, Köhler G, Biedenkapp A, Hutter F. Sample-efficient automated deep reinforcement learning. arXiv. 2020. https://doi.org/10.48550/arXiv.2009.01555

53. Bergstra J, Bengio Y. Random search for hyper-parameter optimization. J Mach Learn Res. 2012;13(2):281–305.

54. Snoek J, Larochelle H, Adams RP. Practical bayesian optimization of machine learning algorithms. Adv Neural Inf Proces Syst. 2012;25.

55. Zahavy T, Xu Z, Veeriah V, Hessel M, Oh J, van Hasselt HP, Silver D, Singh S. A self-tuning actor-critic algorithm. Adv Neural Inf Proces Syst. 2020;33:20913–20924.

56. Eriksson A, Capi G, Doya K. Evolution of meta-parameters in reinforcement learning algorithm. Paper presented at: IEEE: Proceedings of the 2003 IEEE/RSJ International Conference on Intelligent Robots and Systems (IROS 2003); 2003 Oct 27–31; Las Vegas, NV.

57. Elfwing S, Uchibe E, Doya K, Christensen HI. Co-evolution of shaping rewards and meta-parameters in reinforcement learning. Adapt Behav. 2008;16(6):400–412.

58. Jaderberg M, Czarnecki WM, Dunning I, Marris L, Lever G, Castaneda AG, Beattie C, Rabinowitz NC, Morcos AS, Ruderman A, et al. Human-level performance in 3D multiplayer games with population-based reinforcement learning. Science. 2019;364(6443):859–865.

59. Schmitt S, Hudson JJ, Zidek A, Osindero S, Doersch C, Czarnecki WM, Leibo JZ, Kuttler H, Zisserman A, Simonyan K, et al. Kickstarting deep reinforcement learning. arXiv. 2018. https://doi.org/10.48550/arXiv.1803.03835

60. Liu S, Lever G, Merel J, Tunyasuvunakool S, Heess N, Graepel T. Emergent coordination through competition. Paper presented at: International Conference on Learning Representations; 2019 May 6; New Orleans (LA).

61. Wu TR, Wei TH, Wu IC. Accelerating and improving alphazero using population based training. Paper presented at: Proceedings of the AAAI Conference on Artificial Intelligence; 2020 Feb 7–12; New York, NY.

62. Vavak F, Fogarty TC. Comparison of steady state and generational genetic algorithms for use in nonstationary environments. Paper presented at: Proceedings of IEEE International Conference on Evolutionary Computation, IEEE; 1996 May 20–22; Nagoya, Japan.

63. Dalibard V, Jaderberg M. Faster improvement rate population based training. arXiv. 2021. https://arxiv.org/abs/2109.13800

64. Fernandez FC, Caarls W. Parameters tuning and optimization for reinforcement learning algorithms using evolutionary computing. Paper presented at: 2018 International Conference on Information Systems and Computer Science, IEEE; 2018 Nov 13–15; Quito, Ecuador.

65. Cui X, Zhang W, Tüske Z, Picheny M. Evolutionary stochastic gradient descent for optimization of deep neural networks. Paper presented at: Advances in Neural Information Processing Systems; 2018 Dec 2–8; Montréal, Canada.

66. Schneider L, Pfisterer F, Thomas J, Bischl B. A collection of quality diversity optimization problems derived from hyperparameter optimization of machine learning models. Paper presented at: Proceedings of the Genetic and

67. Stanley KO, Clune J, Lehman J, Miikkulainen R. Designing neural networks through neuroevolution. Nat Mach Intell. 2019;1(1):24–35.

68. Gaier A, Ha D. Weight agnostic neural networks. Paper presented at: Thirty-third Conference on Neural Information Processing Systems; 2019 Dec 8–14; Vancouver, Canada.

69. Chrabaszcz P, Loshchilov I, Hutter F. Back to basics: Benchmarking canonical evolution strategies for playing atari. Paper presented at: International Joint Conference on Artificial Intelligence; 2018 Jul 13–19; Stockholm, Sweden.

70. Whiteson S. Evolutionary computation for reinforcement learning. Berlin Heidelberg: Springer; 2012.

71. Choromanski K, Rowland M, Sindhwani V, Turner R, Weller A. Structured evolution with compact architectures for scalable policy optimization. Paper presented at: International Conference on Machine Learning. PMLR; 2018 Jul 10–15; Stockholm, Sweden.

72. Choromanski KM, Pacchiano A, Parker-Holder J, Tang Y, Sindhwani V. From complexity to simplicity: Adaptive ES-active subspaces for blackbox optimization. Paper presented at: Advances in Neural Information Processing Systems; 2019 Dec 8–14; Vancouver, Canada.

73. Tang Y, Choromanski K, Kucukelbir A. Variance reduction for evolution strategies via structured control variates. Paper presented at: International Conference on Artificial Intelligence and Statistics. PMLR. 2020 Aug 26–28; Palermo, Italy.

74. Maheswaranathan N, Metz L, Tucker G, Choi D, Sohl-Dickstein J. Guided evolutionary strategies: Augmenting random search with surrogate gradients. Paper presented at: Proceedings of the 36th International Conference on Machine Learning, PMLR; 2019 Jun 9–15; Long Beach, CA.

75. Liu F-Y, Li Z-N, Qian C. Self-guided evolution strategies with historical estimated gradients. Paper presented at: International Joint Conference on Artifcial Intelligence; 2020 Jan 7–15; Yokohama, Japan.

76. Liu G, Zhao L, Yang F, Bian J, Qin T, Yu N, Liu T-Y. Trust region evolution strategies. Proc AAAI Conf Artif Intell. 2019;33(01):4352–4359.

77. Yi S, Wierstra D, Schaul T, Schmidhuber J. Stochastic search using the natural gradient. Paper presented at: International Conference on Machine Learning; 2009 Jun 14–18; Montreal, Quebec, Canada.

78. Sehnke F, Osendorfer C, Rückstiess T, Graves A, Peters J, Schmidhuber J. Parameter-exploring policy gradients. Neural Netw. 2010;23(4):551–559.

79. Zhang X, Clune J, Stanley KO. On the relationship between the openai evolution strategy and stochastic gradient descent. arXiv. 2017. https://arxiv.org/abs/1712.06564

80. Lehman J, Chen J, Clune J, Stanley KO. ES is more than just a traditional finite-difference approximator. Paper presented at: Proceedings of the Genetic and Evolutionary Computation Conference; 2018 Jul 15–19; Kyoto, Japan.

81. Fuks L, Awad NH, Hutter F, Lindauer M. An evolution strategy with progressive episode lengths for playing games. Paper presented at: International Joint Conferences on Artificial Intelligence; 2019 Aug 10–16; Macao, China.

82. Igel C. Neuroevolution for reinforcement learning using evolution strategies. Paper presented at: The Congress on

67. Stanley KO, Clune J, Lehman J, Miikkulainen R. Designing neural networks through neuroevolution. Nat Mach Intell. 2019;1(1):24–35.

Evolutionary Computation, vol. 4. IEEE; 2003 Dec 8–12; Canberra, ACT, Australia.

83. Heidrich-Meisner V, Igel C. Hoeffding and bernstein races for selecting policies in evolutionary direct policy search. Paper presented at: International Conference on Machine Learning; 2009 Jun 14–18; Montreal, Canada.

84. Heidrich-Meisner V, Igel C. Neuroevolution strategies for episodic reinforcement learning. *J Algorithms*. 2009;64(4):152–168.

85. Chen Z, Zhou Y, He X, Jiang S. A restart-based rank-1 evolution strategy for reinforcement learning. Paper presented at: International Joint Conferences on Artificial Intelligence; 2019 Aug 10–16; Macao, China.

86. Li Z, Zhang Q. A simple yet efficient evolution strategy for large-scale black-box optimization. *IEEE Trans Evol Comput*. 2017;22(5):637–646.

87. Loshchilov I, Glasmachers T, Beyer H-G. Large scale black-box optimization by limited-memory matrix adaptation. *IEEE Trans Evol Comput*. 2018;23(2):353–358.

88. Li Z, Zhang Q, Lin X, Zhen H-L. Fast covariance matrix adaptation for large-scale black-box optimization. *IEEE Trans Cybern*. 2020;50(5):2073–2083.

89. Wieland AP. Evolving controls for unstable systems. In: *Connectionist models*. Morgan Kaufmann Publishers, Inc.; 1991. p. 91–102.

90. Stanley KO, Bryant BD, Miikkulainen R. Evolving adaptive neural networks with and without adaptive synapses. Paper presented at: The 2003 Congress on Evolutionary Computation, vol. 4, IEEE; 2003 Dec 8–12; Canberra, ACT, Australia.

91. Stanley KO, Miikkulainen R. Competitive coevolution through evolutionary complexification. *J Artif Intell Res*. 2004;21:63–100.

92. Stanley KO, Bryant BD, Miikkulainen R. Evolving neural network agents in the nero video game. Paper presented at: Proceedings of the IEEE 2005 Symposium on Computational Intelligence and Games; 2005 Apr 4–6; Essex, UK.

93. Kohl N, Miikkulainen R. Evolving neural networks for strategic decision-making problems. *Neural Netw*. 2009;22(3):326–337.

94. Kassahun Y, Sommer G. Efficient reinforcement learning through evolutionary acquisition of neural topologies. Paper presented at: Proceedings of The European Symposium on Artificial Neural Networks; 2005 Apr 27–29; Bruges, Belgium.

95. Moriguchi H, Honiden S. CMA-TWEANN: Efficient optimization of neural networks via self-adaptation and seamless augmentation. Paper presented at: Proceedings of the 14th Annual Conference on Genetic and Evolutionary Computation; 2012 July 7–11; Philadelphia, PA.

96. Such FP, Madhavan V, Conti E, Lehman J, Stanley KO, Clune J. Deep neuroevolution: Genetic algorithms are a competitive alternative for training deep neural networks for reinforcement learning. arXiv. 2017. https://arxiv.org/abs/1712.06567

97. Le Clei M, Bellec P. Neuroevolution of recurrent architectures on control tasks. Paper presented at: International Conference on Learning Representations Workshop on Agent Learning in Open-Endedness; 2022 Apr 29; Boston, MA.

98. Ha D, Schmidhuber J. Recurrent world models facilitate policy evolution. In: *Advances in neural information processing systems*; 2018. vol. 31.

99. Koutník J, Schmidhuber J, Gomez F. Evolving deep unsupervised convolutional networks for vision-based reinforcement learning. Paper presented at: Proceedings of the 2014 Annual Conference on Genetic and Evolutionary Computation; 2014 Jul 12–16; Vancouver, BC, Canada.

100. Alvernaz S, Togelius J. Autoencoder-augmented neuroevolution for visual doom playing. Paper presented at: 2017 IEEE Conference on Computational Intelligence and Games, IEEE; 2017 Aug 22–25; New York, NY.

101. Risi S, Stanley KO. Deep neuroevolution of recurrent and discrete world models. Paper presented at: Proceedings of the Genetic and Evolutionary Computation Conference; 2019 Jul 13–17; Prague, Czech Republic.

102. Whiteson S, Stone P. Evolutionary function approximation for reinforcement learning. *J Mach Learn Res*. 2006;7(31):877–917.

103. Whiteson S, Stone P. Sample-efficient evolutionary function approximation for reinforcement learning. *Proc Natl Conf Artif Intell*. 2006;21(1):518.

104. Whiteson S, Taylor ME, Stone P. Critical factors in the empirical performance of temporal difference and evolutionary methods for reinforcement learning. *Auton Agent Multi-Agent Syst*. 2010;21(1):1–27.

105. Potter MA, Jong KAD. Cooperative coevolution: An architecture for evolving coadapted subcomponents. *Evol Comput*. 2000;8(1):1–29.

106. Moriarty DE, Mikkulainen R. Efficient reinforcement learning through symbiotic evolution. *Mach Learn*. 1996;22(1):11–32.

107. Gomez F, Miikkulainen R. Solving non-markovian control tasks with neuroevolution. Paper presented at: Proceeding of the Sixteenth International Joint Conference on Artificial Intelligence; 1999 Jul 31–Aug 6; Stockholm, Sweden.

108. Chandra R, Frean M, Zhang M, Omlin CW. Encoding subcomponents in cooperative co-evolutionary recurrent neural networks. *Neurocomputing*. 2011;74(17):3223–3234.

109. Gomez F, Schmidhuber J, Miikkulainen R, Mitchell M. Accelerated neural evolution through cooperatively coevolved synapses. *J Mach Learn Res*. 2008;9(31):937–965.

110. García-Pedrajas N, Hervás-Martínez C, Muñoz-Pérez J. Covnet: A cooperative coevolutionary model for evolving artificial neural networks. *IEEE Trans Neural Netw*. 2003;14(3):575–596.

111. Reisinger J, Stanley KO. Evolving reusable neural modules. In: *Genetic and evolutionary computation conference*. Springer; 2004. p. 69–81.

112. Yang P, Zhang H, Yu Y, Li M, Tang K. Evolutionary reinforcement learning via cooperative coevolutionary negatively correlated search. *Swarm Evol Comput*. 2022;68:Article 100974.

113. Gruau F. Automatic definition of modular neural networks. *Adapt Behav*. 1994;3(2):151–183.

114. Hornby GS, Pollack JB. Creating high-level components with a generative representation for body-brain evolution. *Artif Life*. 2002;8(3):223–246.

115. Stanley KO, Miikkulainen R. A taxonomy for artificial embryogeny. *Artif Life*. 2003;9(2):93–130.

116. Stanley KO. Compositional pattern producing networks: A novel abstraction of development. *Genet Program Evolvable Mach*. 2007;8(2):131–162.

117. Stanley KO, D'Ambrosio DB, Gauci J. A hypercube-based encoding for evolving large-scale neural networks. *Artif Life*. 2009;15(2):185–212.

118. Clune J, Stanley KO, Pennock RT, Ofria C. On the performance of indirect encoding across the continuum of regularity. *IEEE Trans Evol Comput.* 2011;15(3):346–367.

119. Gauci J, Stanley KO. A case study on the critical role of geometric regularity in machine learning. Paper presented at: Proceedings of the 23rd National Conference on Artificial Intelligence, AAAI Press; 2008 Jul 13–17; Chicago, IL.

120. Hausknecht M, Lehman J, Miikkulainen R, Stone P. A neuroevolution approach to general atari game playing. *IEEE Trans Comput Intell AI Games.* 2014;6(4):355–366.

121. Risi S, Stanley KO. Indirectly encoding neural plasticity as a pattern of local rules. In: *International conference on simulation of adaptive behavior.* Springer; 2010. p. 533–543.

122. Risi S, Stanley KO. An enhanced hypercube-based encoding for evolving the placement, density, and connectivity of neurons. *Artif Life.* 2012;18(4):331–363.

123. Deb K, Pratap A, Agarwal S, Meyarivan T. A fast and elitist multiobjective genetic algorithm: NSGA-II. *IEEE Trans Evol Comput.* 2002;6(2):182–197.

124. Huizinga J, Mouret J-B, Clune J. Does aligning phenotypic and genotypic modularity improve the evolution of neural networks? Paper presented at: Proceedings of the Genetic and Evolutionary Computation Conference; 2016 Jul 20–24; Denver, CO.

125. Koutník J, Cuccu G, Schmidhuber J, Gomez F. Evolving large-scale neural networks for vision-based reinforcement learning. Paper presented at: Proceedings of the 15th Annual Conference on Genetic and Evolutionary Computation; 2013 Jul 6–10; Amsterdam, The Netherlands.

126. Clune J, Beckmann BE, Pennock RT, Ofria C. Hybrid: A hybridization of indirect and direct encodings for evolutionary computation. In: *European conference on artificial life.* Springer; 2009. p. 134–141.

127. Vargas-Hákim G-A, Mezura-Montes E, Acosta-Mesa H-G. Hybrid encodings for neuroevolution of convolutional neural networks: A case study. Paper presented at: Proceedings of the Genetic and Evolutionary Computation Conference Companion; 2021 Jul 10–14; Lille, France.

128. Schrum J, Capps B, Steckel K, Volz V, Risi S. Hybrid encoding for generating large scale game level patterns with local variations. *IEEE Trans Games.* 2022;15(1):46–55.

129. Deb K, Kumar A. Real-coded genetic algorithms with simulated binary crossover: Studies on multimodal and multiobjective problems. *Complex Systems.* 1995;9(6):431–454.

130. Gangwani T, Peng J. Genetic policy optimization. Paper presented at: International Conference on Learning Representations; 2018 April 30–May 3; Vancouver, BC, Canada.

131. Bodnar C, Day B, Lió P. Proximal distilled evolutionary reinforcement learning. *Proc AAAI Conf Artif Intell.* 2020;34(04):3283–3290.

132. Franke JK, Köhler G, Awad N, Hutter F. Neural architecture evolution in deep reinforcement learning for continuous control. arXiv. 2019. https://arxiv.org/abs/1910.12824

133. Lehman J, Chen J, Clune J, Stanley KO. Safe mutations for deep and recurrent neural networks through output gradients. Paper presented at: Proceedings of the Genetic and Evolutionary Computation Conference; 2018 Jul 15–19; Kyoto, Japan.

134. Marchesini E, Corsi D, Farinelli A. Exploring safer behaviors for deep reinforcement learning. *Proc AAAI Conf Artif Intell.* 2022;36(7):7701–7709.

135. Uriot T, Izzo D. Safe crossover of neural networks through neuron alignment. Paper presented at: Proceedings of the 2020 Genetic and Evolutionary Computation Conference; 2020 Jul 8-12; Cancún, Mexico.

136. Woodward JR. Evolving turing complete representations. Paper presented at: The Congress on Evolutionary Computation, vol. 2. IEEE; 2003 Dec 8–12; Canberra, ACT, Australia.

137. Miller JF. Cartesian genetic programming. In: *Cartesian genetic programming.* Berlin, Heidelberg: Springer; 2011. p. 17–34.

138. Kelly S, Smith RJ, Heywood MI. Emergent policy discovery for visual reinforcement learning through tangled program graphs: A tutorial. In: *Genetic programming theory and practice XVI.* Ann Arbor (MI): Springer; 2019. p. 37–57.

139. Koza JR, Rice JP. Automatic programming of robots using genetic programming. Paper presented at: Proceedings of the Tenth National Conference on Artificial Intelligence, AAAI Press; 1992 Jul 12–16; San Jose, CA.

140. Ok S, Miyashita K, Hase K. Evolving bipedal locomotion with genetic programming—A preliminary report. Paper presented at: Proceedings of the 2001 Congress on Evolutionary Computation, vol. 2. IEEE; 2001 May 27–30; Seoul, South Korea.

141. Dracopoulos DC, Effraimidis D, Nichols BD. Genetic programming as a solver to challenging reinforcement learning problems. *Int J Comput Res.* 2013;20(3):351–379.

142. Kamio S, Iba H. Adaptation technique for integrating genetic programming and reinforcement learning for real robots. *IEEE Trans Evol Comput.* 2005;9(3):318–333.

143. Gruau F, Whitley D, Pyeatt L. A comparison between cellular encoding and direct encoding for genetic neural networks. Paper presented at: Proceedings of the 1st Annual Conference on Genetic Programming; 1996 Jul 28–31; Stanford, CA.

144. Khan MM, Ahmad AM, Khan GM, Miller JF. Fast learning neural networks using cartesian genetic programming. *Neurocomputing.* 2013;121:274–289.

145. Turner AJ, Miller JF. Neuroevolution: Evolving heterogeneous artificial neural networks. *Evol Intel.* 2014;7(3):135–154.

146. Wilson DG, Cussat-Blanc S, Luga H, Miller JF. Evolving simple programs for playing atari games. Paper presented at: Proceedings of the Genetic and Evolutionary Computation Conference; 2018 Jul 15–19; Kyoto, Japan.

147. Kelly S, Heywood MI. Emergent tangled graph representations for atari game playing agents. In: *European conference on genetic programming.* Springer; 2017. p. 64–79.

148. Kelly S, Heywood MI. Emergent tangled program graphs in multi-task learning. Paper presented at: International Joint Conference on Artificial Intelligence; 2018 Jul 13–19; Stockholm, Sweden.

149. Kelly S, Voegerl T, Banzhaf W, Gondro C. Evolving hierarchical memory-prediction machines in multi-task reinforcement learning. *Genet Program Evolvable Mach.* 2021;22(4):573–605.

150. Smith RJ, Heywood MI. A model of external memory for navigation in partially observable visual reinforcement learning tasks. In: *European conference on genetic programming.* Springer; 2019. p. 162–177.

151. Smith RJ, Heywood MI. Evolving dota 2 shadow fiend bots using genetic programming with external memory. Paper presented at: Proceedings of the Genetic and Evolutionary Computation Conference; 2019 Jul 13–17; Prague, Czech Republic.

152. Onderwater M, Bhulai S, van der Mei R. Value function discovery in markov decision processes with evolutionary algorithms. *IEEE Trans Syst Man Cybern Syst*. 2015;46(9):1190–1201.

153. Hein D, Udluft S, Runkler TA. Interpretable policies for reinforcement learning by genetic programming. *Eng Appl Artif Intell*. 2018;76:158–169.

154. Alibekov E, Kubalík J, Babuška R. Symbolic method for deriving policy in reinforcement learning. Paper presented at: 2016 IEEE 55th Conference on Decision and Control. IEEE; 2016 Dec 12–14; Las Vegas, NV.

155. Derner E, Kubalík J, Babuška R. Data-driven construction of symbolic process models for reinforcement learning. Paper presented at: IEEE International Conference on Robotics and Automation. 2018; 2018 May 21–25; Brisbane, QLD, Australia.

156. Girgin S, Preux P. Feature discovery in reinforcement learning using genetic programming. In: *European conference on genetic programming*. Springer; 2008. p. 218–229.

157. Krawiec K. Genetic programming-based construction of features for machine learning and knowledge discovery tasks. *Genet Program Evolvable Mach*. 2002;3(4):329–343.

158. Plappert M, Houthooft R, Dhariwal P, Sidor S, Chen RY, Chen X, Asfour T, Abbeel P, Andrychowicz M. Parameter space noise for exploration. Paper presented at: International Conference on Learning Representations; 2018 April 30–May 3; Vancouver, BC, Canada.

159. Yang T, Tang H, Bai C, Liu J, Hao J, Meng Z, Liu P, Wang Z. Exploration in deep reinforcement learning: A comprehensive survey. arXiv. 2021. https://arxiv.org/abs/2109.06668v1

160. Pugh JK, Soros LB, Stanley KO. Quality diversity: A new frontier for evolutionary computation. *Front Robot AI*. 2016;3:Article 40.

161. Gravina D, Liapis A, Yannakakis G. Surprise search: Beyond objectives and novelty. Paper presented at: Proceedings of the Genetic and Evolutionary Computation Conference; 2016 Jul 20–24; Denver, CO.

162. Mengistu H, Lehman J, Clune J. Evolvability search: Directly selecting for evolvability in order to study and produce it. Paper presented at: Proceedings of the Genetic and Evolutionary Computation Conference; 2016 Jul 20–24; Denver, CO.

163. Pathak D, Agrawal P, Efros AA, Darrell T. Curiosity-driven exploration by self-supervised prediction. Paper presented at: International Conference on Machine Learning. PMLR; 2017 Aug 6–11; Sydney, NSW, Australia.

164. Risi S, Vanderbleek SD, Hughes CE, Stanley KO. How novelty search escapes the deceptive trap of learning. Paper presented at: Proceedings of the 11th Annual Conference on Genetic and Evolutionary Computation; 2009 Jul 8–12; Montreal, Québec, Canada.

165. Cuccu G, Gomez F. When novelty is not enough. In: *European conference on the applications of evolutionary computation*. Springer; 2011. p. 234–243.

166. Mouret J-B, Doncieux S. Encouraging behavioral diversity in evolutionary robotics: An empirical study. *Evol Comput*. 2012;20(1):91–133.

167. Lehman J, Stanley KO. Evolving a diversity of virtual creatures through novelty search and local competition. Paper presented at: Proceedings of the 13th Annual Conference on Genetic and Evolutionary Computation; 2011 Jul 12–16; Dublin, Ireland.

168. Liu Q, Wang Y, Liu X. PNS: Population-guided novelty search for reinforcement learning in hard exploration environments. Paper presented at: 2021 IEEE/RSJ International Conference on Intelligent Robots and Systems; 2021 Sep 27–Oct 1; Prague, Czech Republic.

169. Mouret J-B, Clune J. Illuminating search spaces by mapping elites. arXiv. 2015. https://arxiv.org/abs/1504.04909

170. Cully A. Autonomous skill discovery with quality-diversity and unsupervised descriptors. Paper presented at: Proceedings of the Genetic and Evolutionary Computation Conference; 2019 Jul 1–17; Prague, Czech Republic.

171. Tao RY, François-Lavet V, Pineau J. Novelty search in representational space for sample efficient exploration. *Adv Neural Inf Proces Syst*. 2020;33:8114–8126.

172. Rakicevic N, Cully A, Kormushev P. Policy manifold search: Exploring the manifold hypothesis for diversity-based neuroevolution. Paper presented at: Genetic and Evolutionary Computation Conference; 2021 Jul 10–14; Lille, France.

173. Parker-Holder J, Pacchiano A, Choromanski K, Roberts S. Effective diversity in population-based reinforcement learning. arXiv. 2020. https://arxiv.org/abs/2002.00632v1

174. Jackson EC, Daley M. Novelty search for deep reinforcement learning policy network weights by action sequence edit metric distance. Paper presented at: Proceedings of the Genetic and Evolutionary Computation Conference Companion; 2019 Jul 13–17; Prague, Czech Republic.

175. Keller L, Tanneberg D, Stark S, Peters J. Model-based quality-diversity search for efficient robot learning. Paper presented at: 2020 IEEE/RSJ International Conference on Intelligent Robots and Systems. IEEE; 2020 Oct 24–2021 Jan 24; Las Vegas, NV.

176. Salehi A, Coninx A, Doncieux S. Few-shot quality-diversity optimization. *IEEE Robot Autom Lett*. 2022;7(2):4424–4431.

177. Wang Y, Xue K, Qian C. Evolutionary diversity optimization with clustering-based selection for reinforcement learning. Paper presented at: International Conference on Learning Representations; 2022 Apr 25–29; Virtual conference.

178. Wang R, Lehman J, Clune J, Stanley KO. Poet: Open-ended coevolution of environments and their optimized solutions. Paper presented at: Proceedings of the Genetic and Evolutionary Computation Conference; 2019 Jul 13–17; Prague, Czech Republic.

179. Wang R, Lehman J, Rawal A, Zhi J, Li Y, Clune J, Stanley K. Enhanced POET: Open-ended reinforcement learning through unbounded invention of learning challenges and their solutions. Paper presented at: International Conference on Machine Learning. PMLR; 2020 Jul 12–18; Virtual conference.

180. Bhatt V, Tjanaka B, Fontaine MC, Nikolaidis S. Deep surrogate assisted generation of environments. arXiv. 2022. https://arxiv.org/abs/2206.04199

181. Brych S, Cully A. Competitiveness of map-elites against proximal policy optimization on locomotion tasks in deterministic simulations. arXiv. 2020. https://arxiv.org/abs/2009.08438

182. Vassiliades V, Chatzilygeroudis K, Mouret J-B. Using centroidal voronoi tessellations to scale up the multidimensional archive of phenotypic elites algorithm. *IEEE Trans Evol Comput*. 2017;22(4):623–630.

183. Colas C, Huizinga J, Madhavan V, Clune J. Scaling map-elites to deep neuroevolution. arXiv. 2020. https://arxiv.org/abs/2003.01825

184. Pierrot T, Macé V, Chalumeau F, Flajolet A, Cideron G, Beguir K, Cully A, Sigaud O, Perrin-Gilbert N. Diversity policy gradient for sample efficient quality-diversity optimization. Paper presented at: ICLR Workshop on Agent Learning in Open-Endedness; 2022 Apr 25–29; Boston, MA.

185. Tjanaka B, Fontaine MC, Togelius J, Nikolaidis S. Differentiable quality diversity for reinforcement learning by approximating gradients. Paper presented at: International Conference on Learning Representations Workshop on Agent Learning in Open-Endedness; 2022 Apr 25–29; Boston, MA.

186. Nilsson O, Cully A. Policy gradient assisted map-elites. Paper presented at: Genetic and Evolutionary Computation Conference; 2021 Jul 10–14; Lille, France.

187. Zhang Y, Fontaine MC, Hoover AK, Nikolaidis S. Dsa-me: Deep surrogate assisted map-elites. Paper presented at: International Conference on Learning Representations Workshop on Agent Learning in Open-Endedness; 2022 Apr 25–29; Boston, MA.

188. Ecoffet A, Huizinga J, Lehman J, Stanley KO, Clune J. First return, then explore. *Nature*. 2021;590(7847):580–586.

189. Gravina D, Liapis A, Yannakakis GN. Quality diversity through surprise. *IEEE Trans Evol Comput*. 2018;23(4):603–616.

190. Bellemare M, Srinivasan S, Ostrovski G, Schaul T, Saxton D, Munos R. Unifying count-based exploration and intrinsic motivation. *Adv Neural Inf Proces Syst*. 2016;29:1471–1479.

191. Forestier S, Portelas R, Mollard Y, Oudeyer P.-Y. Intrinsically motivated goal exploration processes with automatic curriculum learning. arXiv. 2017. https://arxiv.org/abs/1708.02190

192. Colas C, Sigaud O, Oudeyer P-Y. GEP-PG: Decoupling exploration and exploitation in deep reinforcement learning algorithms. Paper presented at: International Conference on Machine Learning. PMLR; 2018 Jul 10–15; Stockholm, Sweden.

193. Stanton C, Clune J. Deep curiosity search: Intra-life exploration improves performance on challenging deep reinforcement learning problems. arXiv. 2018. https://arxiv.org/abs/1806.00553

194. HAO J, Li P, Tang H, ZHENG Y, Fu X, Meng Z. ERL-Re$^2$: Efficient evolutionary reinforcement learning with shared state representation and individual policy representation. Paper presented at: The Eleventh International Conference on Learning Representations; 2023 May 1–5; Kigali, Rwanda.

195. Zheng H, Jiang J, Wei P, Long G, Zhang C. Competitive and cooperative heterogeneous deep reinforcement learning. Paper presented at: Proceedings of the International Joint Conference on Autonomous Agents and Multiagent Systems, 2020 May; Auckland, New Zealand.

196. Lü S, Han S, Zhou W, Zhang J. Recruitment-imitation mechanism for evolutionary reinforcement learning. *Inf Sci*. 2021;553:172–188.

197. Ma Y, Liu T, Wei B, Liu Y, Xu K, Li W. Evolutionary action selection for gradient-based policy learning. arXiv. 2022. https://arxiv.org/abs/2201.04286

198. Shi H, Zhou B, Zeng H, Wang F, Dong Y, Li J, Wang K, Tian H, Meng MQ-H. Reinforcement learning with evolutionary trajectory generator: A general approach for quadrupedal locomotion. *IEEE Robot Autom Lett*. 2022;7(2):3085–3092.

199. Morel A, Kunimoto Y, Coninx A, Doncieux S. Automatic acquisition of a repertoire of diverse grasping trajectories through behavior shaping and novelty search. arXiv. 2022. http://arxiv.org/abs/2205.08189

200. Pourchot A, Sigaud O. Cem-rl: Combining evolutionary and gradient-based methods for policy search. Paper presented at: International Conference on Learning Representations; 2019 May 6–9; New Orleans, LA.

201. Lee K, Lee B-U, Shin U, Kweon IS. An efficient asynchronous method for integrating evolutionary and gradient-based policy search. *Adv Neural Inf Proces Syst*. 2020;33:10 124–10 135.

202. Suri K. Off-policy evolutionary reinforcement learning with maximum mutations. Paper presented at: Proceedings of the 21st International Conference on Autonomous Agents and Multiagent Systems; 2022 May 9–13; Virtual Event, New Zealand.

203. Marchesini E, Corsi D, Farinelli A. Genetic soft updates for policy evolution in deep reinforcement learning. Paper presented at: International Conference on Learning Representations; 2020 Apr 30; Addis Ababa, Ethiopia.

204. Zhu S, Belardinelli F, León BG. Evolutionary reinforcement learning for sparse rewards. Paper presented at: Proceedings of the Genetic and Evolutionary Computation Conference; 2021 Jul 10–14; Lille, France.

205. Clune J. AI-GAs: Ai-generating algorithms, an alternate paradigm for producing general artificial intelligence. arXiv. 2019. https://arxiv.org/abs/1905.10985

206. Faust A, Francis A, Mehta D. Evolving rewards to automate reinforcement learning. arXiv. 2019. https://arxiv.org/abs/1905.07628

207. Laud A, DeJong G. The influence of reward on the speed of reinforcement learning: An analysis of shaping. Paper presented at: Proceedings of the 20th International Conference on Machine Learning; 2003 Aug 21–24; Washington, DC.

208. Ng AY, Harada D, Russell S. Policy invariance under reward transformations: Theory and application to reward shaping. Paper presented at: International Conference on Machine Learning; 1999 Jun 27–30; Bled, Slovenia.

209. Ferreira F, Nierhoff T, Saelinger A, Hutter F. Learning synthetic environments and reward networks for reinforcement learning. Paper presented at: International Conference on Learning Representations; 2022 Apr 25–29; Virtual conference.

210. Singh S, Lewis RL, Barto AG, Sorg J. Intrinsically motivated reinforcement learning: An evolutionary perspective. *IEEE Trans Auton Ment Dev*. 2010;2(2):70–82.

211. Niekum S, Barto AG, Spector L. Genetic programming for reward function search. *IEEE Trans Auton Ment Dev*. 2010;2(2):83–90.

212. Uchibe E, Doya K. Finding intrinsic rewards by embodied evolution and constrained reinforcement learning. *Neural Netw*. 2008;21(10):1447–1455.

213. Sheikh HU, Khadka S, Miret S, Majumdar S, Phielipp M. Learning intrinsic symbolic rewards in reinforcement learning. Paper presented at: International Joint Conference on Neural Networks. IEEE; 2022 Jul 18–23; Padua, Italy.

214. Paolo G, Coninx A, Doncieux S, Laflaquière A. Sparse reward exploration via novelty search and emitters. Paper presented at: Proceedings of the Genetic and Evolutionary Computation Conference, 2021 Jul 10–14; Lille, France.

215. Majumdar S, Khadka S, Miret S, Mcaleer S, Tumer K. Evolutionary reinforcement learning for sample-efficient

multiagent coordination. *Inter Conf Mach Learn*. 2020; Article 617.

216. Lowe R, Wu Y, Tamar A, Harb J, Abbeel P, Mordatch I. Multi-agent actor-critic for mixed cooperative-competitive environments. arXiv. 2017. https://arxiv.org/abs/1706.02275

217. Sachdeva E, Khadka S, Majumdar S, Tumer K. Maedys: Multiagent evolution via dynamic skill selection. Paper presented at: Proceedings of the Genetic and Evolutionary Computation Conference; 2021 Jul 10–14; Lille, France.

218. Chiang H-TL, Faust A, Fiser M, Francis A. Learning navigation behaviors end-to-end with autorl. *IEEE Robot Autom Lett*. 2019;4(2):2007–2014.

219. Wang JX, Hughes E, Fernando C, Czarnecki WM, Duéñez-Guzmán EA, Leibo JZ. Evolving intrinsic motivations for altruistic behavior. arXiv. 2018. https://arxiv.org/abs/1811.05931

220. Finn C, Abbeel P, Levine S. Model-agnostic meta-learning for fast adaptation of deep networks. Paper presented at: International Conference on Machine Learning. JMLR. org; 2017 Aug 6–11; Sydney, Australia.

221. Duan Y, Schulman J, Chen X, Bartlett PL, Sutskever I, Abbeel P. RI²: Fast reinforcement learning via slow reinforcement learning. arXiv. 2016. https://arxiv.org/abs/1611.02779

222. Houthooft R, Chen Y, Isola P, Stadie B, Wolski F, Jonathan Ho O, Abbeel P. Evolved policy gradients. *Adv Neural Inf Proces Syst*. 2018;31:5405–5414.

223. Song X, Gao W, Yang Y, Choromanski K, Pacchiano A, Tang Y. Es-maml: Simple hessian-free meta learning. arXiv. 2019. https://arxiv.org/abs/1910.01215

224. Fernando C, Sygnowski J, Osindero S, Wang J, Schaul T, Teplyashin D, Sprechmann P, Pritzel A, Rusu A. Meta-learning by the baldwin effect. Paper presented at: Proceedings of the Genetic and Evolutionary Computation Conference Companion; 2018 Jul 15–19; Kyoto Japan.

225. Co-Reyes JD, Miao Y, Peng D, Real E, Le QV, Levine S, Lee H, Faust A. Evolving reinforcement learning algorithms. Paper presented at: International Conference on Learning Representations; 2021 May 4; Vienna, Austria.

226. Garau-Luis JJ, Miao Y, Co-Reyes JD, Parisi A, Tan J, Real E, Faust A. Multi-objective evolution for generalizable policy gradient algorithms. Paper presented at: International Conference on Learning Representations; 2022 May 4; Virtual.

227. Alet F, Schneider MF, Lozano-Perez T, Kaelbling LP, Meta-learning curiosity algorithms. Paper presented at: International Conference on Learning Representations; 2020 Apr 30; Addis Ababa, Ethiopia.

228. Coello Coello CA, González Brambila S, Figueroa Gamboa J, Castillo Tapia MG, Hernández Gómez R. Evolutionary multiobjective optimization: open research areas and some challenges lying ahead. *Complex Intell Syst*. 2020;6:221–236.

229. Van Moffaert K, Drugan MM, Nowé A. Scalarized multi-objective reinforcement learning: Novel design techniques. Paper presented at: 2013 IEEE Symposium on Adaptive Dynamic Programming and Reinforcement Learning; 2013 Apr 16–19; Singapore.

230. Bader JM. Hypervolume-based search for multiobjective optimization: theory and methods [thesis]. ETH Zurich; 2010.

231. Zitzler E, Thiele L, Laumanns M, Fonseca CM, da Fonseca VG. Performance assessment of multiobjective optimizers: An analysis and review. *IEEE Trans Evol Comput*. 2003;7(2):117–132.

232. Fonseca CM, Fleming PJ. An overview of evolutionary algorithms in multiobjective optimization. *Evol Comput*. 1995;3(1):1–16.

233. Beume N, Fonseca CM, Lopez-Ibanez M, Paquete L, Vahrenhold J. On the complexity of computing the hypervolume indicator. *IEEE Trans Evol Comput*. 2009;13(5):1075–1082.

234. Xu J, Tian Y, Ma P, Rus D, Sueda S, Matusik W. Prediction-guided multi-objective reinforcement learning for continuous robot control. Paper presented at: International Conference on Machine Learning; 2020 Jul 12–18; Virtual.

235. Feinberg EA, Shwartz A. Constrained markov decision models with weighted discounted rewards. *Math Oper Res*. 1995;20(2):302–320.

236. Abels A, Roijers D, Lenaerts T, Nowé A, Steckelmacher D. Dynamic weights in multi-objective deep reinforcement learning. Paper presented at: International Conference on Machine Learning; 2019 Jun 10–15; Long Beach, CA.

237. Moffaert KV, Drugan MM, Nowé A. Hypervolume-based multi-objective reinforcement learning. In: *International Conference on Evolutionary Multi-Criterion Optimization*. Springer; 2013. pp. 352–366.

238. Yamamoto H, Hayashida T, Nishizaki I, Sekizaki S. Hypervolume-based multi-objective reinforcement learning: Interactive approach. *Adv Sci Technol Eng Syst J*. 2019;4(1):93–100.

239. Van Moffaert K, Nowé A. Multi-objective reinforcement learning using sets of pareto dominating policies. *J Mach Learn Res*. 2014;15(1):3483–3512.

240. Brys T, Harutyunyan A, Vrancx P, Taylor ME, Kudenko D, Nowé A. Multi-objectivization of reinforcement learning problems by reward shaping. Paper presented at: 2014 International Joint Conference on Neural Networks; 2014 Jul 6–11; Beijing, China.

241. Shen R, Zheng Y, Hao J, Meng Z, Chen Y, Fan C, Liu Y, Generating behavior-diverse game AIs with evolutionary multi-objective deep reinforcement learning. Paper presented at: 2020 International Joint Conference on Artificial Intelligence; 2021 Jan 7–15; Yokohama, Japan.

242. Villin V, Masuyama N, Nojima Y, Effects of different optimization formulations in evolutionary reinforcement learning on diverse behavior generation. Paper presented at: IEEE Symposium Series on Computational Intelligence; 2021 Dec 5–7; Orlando, FL.

243. Li B, Li J, Tang K, Yao X. Many-objective evolutionary algorithms: A survey. *ACM Comput Surv*. 2015;48(1):1–35.

244. Han S, Sung Y, Dimension-wise importance sampling weight clipping for sample-efficient reinforcement learning. Paper presented at: International Conference on Machine Learning; 2019 Jun 10–15; Long Beach, CA.

245. Storn R, Price K. Differential evolution—A simple and efficient heuristic for global optimization over continuous spaces. *J Glob Optim*. 1997;11(4):341–359.

246. Kennedy J, Eberhart R. Particle swarm optimization. Paper presented at: Proceedings of International Conference on Neural Networks; 1995 Nov 27–Dec 01; Perth, WA, Australia

247. Cheng R, Jin Y. A competitive swarm optimizer for large scale optimization. *IEEE Trans Cybern*. 2015;45(2):191–204.

248. Stork J, Zaefferer M, Eisler N, Tichelmann P, Bartz-Beielstein T, Eiben A. Behavior-based neuroevolutionary training in

reinforcement learning. Paper presented at: Proceedings of the Genetic and Evolutionary Computation Conference; 2021 Jul 10–14; Lille, France.

249. Wang Y, Zhang T, Chang Y, Wang X, Liang B, Yuan B. A surrogate-assisted controller for expensive evolutionary reinforcement learning. *Inf Sci*. 2022;616:539–557.

250. Brockman G, Cheung V, Pettersson L, Schneider J, Schulman J, Tang J, Zaremba W. OpenAI Gym. arXiv. 2016. https://arxiv.org/abs/1606.01540

251. Bai H, Shen R, Lin Y, Xu B, Cheng R. Lamarckian platform: Pushing the boundaries of evolutionary reinforcement learning towards asynchronous commercial games. *IEEE Trans Games*. 2022.

252. Tangri R, Mandic DP, Constantinides AG, Pearl: Parallel evolutionary and reinforcement learning library. arXiv. 2022. https://arxiv.org/abs/2201.09568

253. Tang Y, Tian Y, Ha D. Evojax: Hardware-accelerated neuroevolution. Paper presented at: Proceedings of the Genetic and Evolutionary Computation Conference Companion; 2022 Jul 9; Boston, MA.

254. Lange RT. Evosax: Jax-based evolution strategies. arXiv. 2022. arXiv:2212.04180.

255. Huang B, Cheng R, Jin Y, Tan KC, Evox: A distributed gpu-accelerated library towards scalable evolutionary computation. arXiv. 2023. https://arxiv.org/abs/2301.12457

256. Lim B, Allard M, Grillotti L, Cully A, Accelerated quality-diversity for robotics through massive parallelism. Paper presented at: ICLR Workshop on Agent Learning in Open-Endedness; 2022 Apr 28; Virtual.

257. Bhatia J, Jackson H, Tian Y, Xu J, Matusik W. Evolution gym: A large-scale benchmark for evolving soft robots. *Adv Neural Inf Proces Syst*. 2021;34:2201–2214.